\documentclass{article} 
\usepackage{iclr2015,times}
\usepackage[breaklinks,colorlinks]{hyperref}
\usepackage{url}
\usepackage{multirow}
\usepackage{amsmath}
\usepackage{amssymb}
\usepackage{color}
\usepackage{graphicx}
\usepackage{caption,subcaption}
\usepackage{multirow}
\usepackage{booktabs}
\definecolor{myblue}{rgb}{.0,.0,.7}
\definecolor{mygreen}{rgb}{.0,.7,.0}
\definecolor{mybrown}{rgb}{.7,.0,.0}
\hypersetup{colorlinks,linkcolor=mygreen,urlcolor=myblue,citecolor=mybrown}

\newcommand{\imw}{1cm}

\newcommand{\mycomment}[1]{} 
\newcommand{\mytilde}{\raise.17ex\hbox{$\scriptstyle\mathtt{\sim}$}}
\newcommand{\comment}[1]{}

\let\citenop\cite
\let\cite\citep

\makeatletter
\newcommand*\mysize{%
  \@setfontsize\mysize{16.4}{0}%
}
\makeatother

\title{\vspace{-5mm}{\mysize Fracking Deep Convolutional Image Descriptors}\vspace{-1mm}}

\author{Edgar Simo-Serra$^*$, Eduard Trulls\thanks{Both first authors contributed equally.} \\
Institut de Rob\`otica i Inform\`atica Industrial (CSIC-UPC). Barcelona, Spain \\
\texttt{\{esimo,etrulls\}@iri.upc.edu}\vspace{-5mm}
\And
Luis Ferraz \\
Catchoom Technologies S.L. Barcelona, Spain \\
\texttt{luis.ferraz@catchoom.com}\vspace{-5mm}
\And
Iasonas Kokkinos \\
CentraleSupelec. Chatenay-Malabry, France \\
\texttt{iasonas.kokkinos@ecp.fr}\vspace{-5mm}
\AND
Francesc Moreno-Noguer \\
Institut de Rob\`otica i Inform\`atica Industrial (CSIC-UPC). Barcelona, Spain \\
\texttt{fmoreno@iri.upc.edu}\vspace{-5mm}
}

\iclrconference 

\begin{document}

\maketitle

\begin{abstract}
\vspace{-2mm}
In this paper we propose a novel framework for learning local image descriptors in a discriminative manner. For this purpose we explore a siamese architecture of Deep Convolutional Neural Networks (CNN), with a Hinge embedding loss on the L$_2$ distance between descriptors. Since a siamese architecture uses pairs rather than single image patches to train, there exist a large number of positive samples and an exponential number of negative samples. We propose to explore this space with a stochastic sampling of the training set, in combination with an aggressive mining strategy over both the positive and negative samples which we denote as ``fracking''. We perform a thorough evaluation of the architecture hyper-parameters, and demonstrate large performance gains compared to both standard CNN learning strategies, hand-crafted image descriptors like SIFT, and the state-of-the-art on learned descriptors: up to 2.5x vs SIFT and 1.5x vs the state-of-the-art in terms of the area under the curve (AUC) of the Precision-Recall curve.
\vspace{-2mm}
\end{abstract}




\section{Introduction}
\label{sec:intro}

Feature descriptors, i.e. the invariant and discriminative representation of local image patches, is a major research topic in computer vision. The field reached maturity with SIFT~\cite{LoweIJCV04}, and has since become the cornerstone of a wide range of applications in recognition and registration. While most descriptors use hand-crafted features~\cite{LoweIJCV04,BayECCV2006,KokkinosTR2012,TrullsCVPR2013,SimoIJCV2015}, there has recently been interest in using machine learning algorithms to learn descriptors from large databases. 

In this paper we draw inspiration on the recent success of Deep Convolutional Neural Networks on large-scale image classification problems~\cite{KrizhevskyNIPS12,SzegedyNIPS13} to build discriminative descriptors for local patches. Specifically, we propose an architecture based on a siamese structure of two CNNs that share the parameters. We compute the L$_2$ norm on their output, i.e. the descriptors, and use a loss that enforces the norm to be small for corresponding patches and large otherwise. We demonstrate that this approach allows us to learn compact and discriminative representations.

To implement this approach we rely on the dataset of~\citenop{BrownPAMI2011}, which contains over 1.5M grayscale $64 \times 64$ image patches from different views of 500K different 3D points. With such large datasets it becomes intractable to exhaustively explore all corresponding and non-corresponding pairs. Random sampling is typically used; however, most correspondences are not useful and hinder the learning of a discriminant mapping. We address this issue with aggressive mining of ``hard" positives and negatives, a strategy that we denote as ``fracking'', and which proves fundamental in order to obtain discriminative learned descriptors. In particular, in some of the tests we obtain up to a 169\% increase in performance with SIFT as a baseline.


\section{Related Work}
\label{sec:related}

Local features have proven very successful at matching points across images, and are nearly ubiquitous in modern computer vision, with a broad range of applications encompassing stereo, pose estimation, classification, detection, medical imaging and many others. Recent developments in the design of local image descriptors are moving from carefully-engineered features~\cite{LoweIJCV04,BayECCV2006} towards learning features from large volumes of data. This line of works includes unsupervised techniques based on hashing as well as supervised approaches using Linear Discriminant Analysis~\cite{BrownPAMI2011,GongPAMI2012,StrechaPAMI2012}, boosting~\cite{TrzcinskiCVPR2013}, and convex optimization~\cite{SimonyanPAMI14}.

In this paper we explore solutions based on deep convolutional networks (CNNs). CNNs have been used in computer vision for decades, but are currenly experiencing a resurgence kickstarted by the accomplishments of \citenop{KrizhevskyNIPS12} on large-scale image classification. The application of CNNs  to the problem of descriptor learning has already been explored by some researchers~\cite{JahrerCVWW2008,OsendorferICONIP13}. These works are however preliminary, and many open questions remain regarding the practical application of CNNs for learning descriptors, such as the most adequate network architectures and application-dependent training schemes. In this paper we aim to provide a rigorous analysis of several of these topics. In particular, we use a siamese network~\cite{BromleyNIPS94} to train the models, and experiment with different network configurations inspired by the state-of-the-art in deep learning.

Additionally, we demonstrate that aggressive mining of both ``hard'' positive and negative matching pairs greatly enhances the learning process. Mining hard negatives is a well-known procedure in sliding-window detectors~\cite{FelzenszwalbPAMI10}, where the number of negative samples is virtually unlimited and yet most negatives are easily discriminated. Similar techniques have been applied to CNNs for object detection~\cite{SzegedyNIPS13,GirshickCVPR14}.



\newcommand{\bx}{\mathbf{x}}

\section{Learning Deep Descriptors}
\label{sec:method}

Given an  intensity patch $\bx \in \mathbb{R}^N$, the descriptor of $\bx$ is a non-linear mapping $D(\bx)$ that is expected to be discriminative, i.e. descriptors for image patches corresponding to the same point should be similar, and dissimilar otherwise.

In the context of multiple-view geometry, descriptors are typically computed for salient points where scale and orientation can be reliably estimated, for invariance. Patches then capture local projections of 3D scenes. Let us consider that each image patch $\bx_i$ has an index $p_i$ that uniquely identifies the 3D point which roughly projects onto the 2D patch, from a specific viewpoint. Therefore, taking the L$_2$ norm as a similarity metric between descriptors, for an ideal descriptor we would wish that
\begin{equation}
d_D(\bx_1,\bx_2) = \| D(\bx_1) - D(\bx_2) \|_2 = \left\{ \begin{array}{l l}
   0    & \quad \text{if $p_1=p_2$} \\
   \infty & \quad \text{if $p_1\neq p_2$} \\
\end{array} \right.
\label{eq:ideal}
\end{equation}

\begin{figure}[t!]
\begin{minipage}[b]{0.32\linewidth}
	\includegraphics[width=\linewidth]{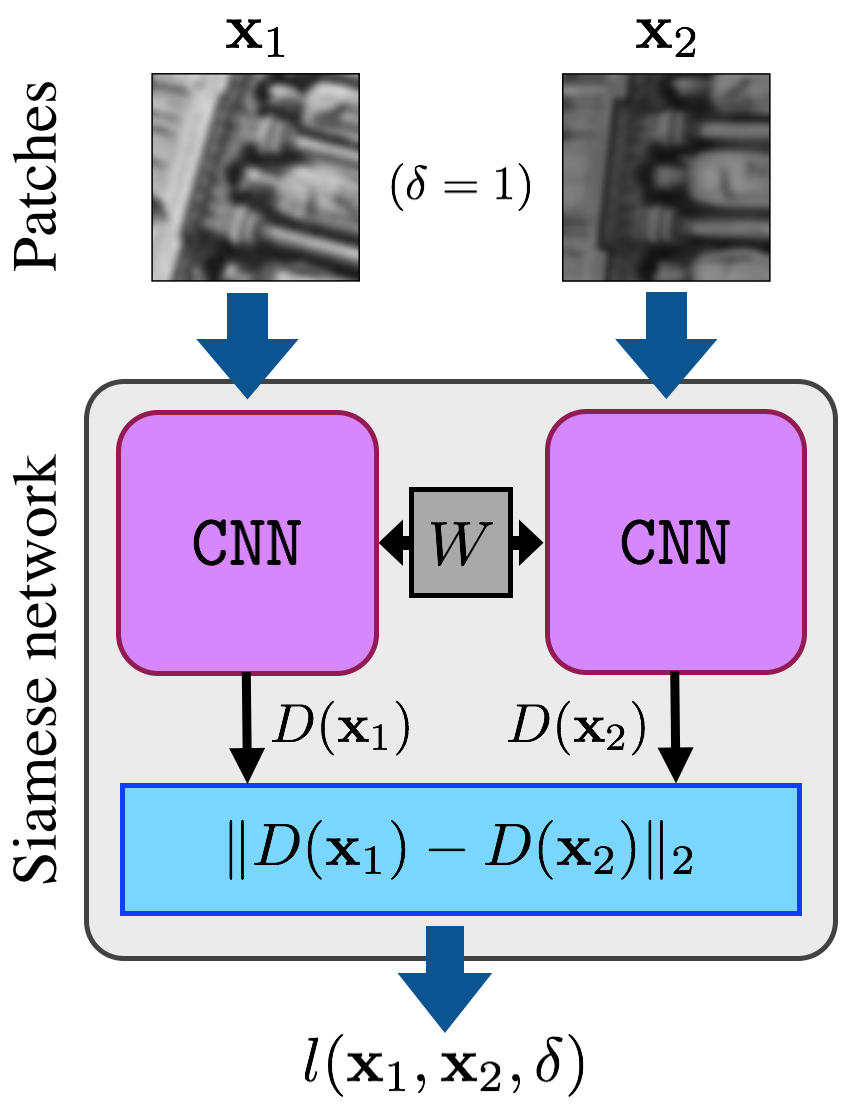}
	\subcaption{Siamese network.}
	\label{fig:arch-siam:siamese}
\end{minipage}
\begin{minipage}[b]{0.68\linewidth}
	\centering
	\renewcommand{\arraystretch}{1.2}
	{\small
		\begin{tabular}{rcccc}
		Name & Layer 1 & Layer 2 & Layer 3 & Layer 4 \\
		\hline \hline
		\multirow{2}{*}{CNN3\_NN1} &  32x7x7 &  64x6x6 & 128x5x5 & 128 \\
		                          & x2 pool & x3 pool & x4 pool &  -  \\
		\hline
		\multirow{2}{*}{CNN3}       &  32x7x7 &  64x6x6 & 128x5x5 &  -  \\
		                            & x2 pool & x3 pool & x4 pool &  -  \\
		\hline
		\multirow{2}{*}{CNN2a\_NN1} &  32x5x5 &  64x5x5 &   128   &  -  \\
		                            & x3 pool & x4 pool &    -    &  -  \\
		\hline
		\multirow{2}{*}{CNN2b\_NN1} &  32x9x9 &  64x5x5 &   128   &  -  \\
		                            & x4 pool & x5 pool &    -    &  -  \\
		\hline
		\multirow{2}{*}{CNN2}       &  64x5x5 & 128x5x5 &    -    &  -  \\
		                            & x4 pool & x11 pool&    -    &  -  \\
		\hline
		\multirow{2}{*}{CNN1\_NN1}  &  32x9x9 &   128   &    -    &  -  \\
		                            & x14 pool&    -    &    -    &  -  \\
		\end{tabular}
	}
	\subcaption{Various convolutional neural network architectures.}
	\label{fig:arch-siam:architectures}
\end{minipage}
\caption{{\bf Left:} Schematic of a siamese network, where pairs of input patches are processed by two copies of the same CNN. {\bf Right:} Different CNN configurations evaluated in this paper.} 
\label{fig:arch-siam}
\vspace{-3mm}
\end{figure}

We propose learning descriptors using a siamese network~\cite{BromleyNIPS94}, i.e. optimizing the model for pairs of corresponding or non-corresponding patches, 
as shown in Fig.~\ref{fig:arch-siam}-(\subref{fig:arch-siam:siamese}). We propagate the patches through the model to extract the descriptors and then compute their L$_2$ norm, which is a standard similarity measure for image descriptors. We then compute the loss function on this distance. Given a pair of  patches $\bx_1$ and $\bx_2$ we  define a loss function of the form
\begin{equation}
l(\bx_1,\bx_2,\delta) = \delta\cdot l_P(d_D(\bx_1,\bx_2)) + (1- \delta) \cdot l_N(d_D(\bx_1,\bx_2))
\end{equation}
\noindent where $\delta$ is the indicator function, which is $1$ if $p_1=p_2$,   and $0$ otherwise. $l_P$ and $l_N$ are the partial loss functions for patches corresponding to the same 3D point and to different points, respectively. When performing back-propagation, the gradients are independently accumulated for both descriptors, but jointly applied to the weights, as they are shared.

Although it would be ideal to optimize directly for Eq.~\eqref{eq:ideal}, we relax it, using a margin $m$ for $l_N(\cdot)$. In particular, we consider the hinge embedding criterion~\cite{MobahiICML2009}
\begin{equation}
l_P(d_D(\bx_1,\bx_2)) = d_D(\bx_1,\bx_2) \qquad \text{and} \qquad l_N(d_D(\bx_1,\bx_2)) = \max(0, m-d_D(\bx_1,\bx_2))
\end{equation}


\subsection{Convolutional Neural Network Descriptors}

When designing the structure of the CNN we are limited by the size of the input data, in our case 64$\times$64 patches from the dataset of~\citenop{BrownPAMI2011}. Note that larger patches would allow us to consider deeper networks, and possibly more informative descriptors, but at the same time they would be also more susceptible to occlusions. We consider networks of up to three convolutional layers, followed by up to a single additional fully-connected layer. We target descriptors of size 128, the same as SIFT~\cite{LoweIJCV04}; this value also constrains the architectures we can explore.

\comment{ 
We consider convolutional neural networks as a way to obtain the non-linear mapping $D(\bx)$. We are limited by the size of the patch, i.e. 64$\times$64 for the dataset of~\citenop{BrownPAMI2011}, which limits the depth of the network; larger patches can be more informative but are also more susceptible to occlusions. We consider networks of up to three convolutional layers, and up to a single additional fully connected layer. We target descriptors of 128 dimensions, the same size as SIFT~\cite{LoweIJCV04}; this value also constrains the architectures we can explore.}


As usual, each convolutional layer consists four sub-layers: filter layer, non-linearity layer, pooling layer and normalization layer. Since  sparser connectivity has been shown to improve performance while lowering parameters and increasing speed~\cite{CulurcielloARXIV13},  except for the first layer, the filters are not densely connected to the previous layers. Instead, they are sparsely connected at random, so that the mean number of connections each input layer has is constant.

\comment{
As usual, each convolutional layer consists four sub-layers: filter layer, non-linearity layer, pooling layer and normalization layer. Except for the first layer, the filters are not densely connected to the previous layers. Instead, they are sparsely connected at random, so that the mean number of connections each input layer has is constant.  as sparser connectivity has been shown to improve performance while lowering parameters and increasing speed~\cite{CulurcielloARXIV13}.}

Regarding the non-linear layer,  we use hyperbolic tangent (Tanh), as we found it performs better than Rectified Linear Units (ReLU). We use L$_2$ pooling for the pooling sublayers, which were shown to outperfom the more standard max pooling~\cite{SermanetICPR12}. Normalization has been shown to be important for deep networks~\cite{JarrettICCV09} and fundamental for descriptors~\cite{MikolajczykPAMI05}. We use subtractive normalization for a 5$\times$5 neighbourhood with a Gaussian kernel. We will justify these decisions empirically in Sec.~\ref{sec:results}.

An overview of the architectures we consider is given in Fig.~\ref{fig:arch-siam}-(\subref{fig:arch-siam:architectures}). We choose a set of six networks, from 2 up to 4 layers. The architecture hyperparameters (number of layers and  convolutional/pooling filter size
) are chosen so that no padding is needed. We consider models with a final fully-connected layer as well as fully convolutional models, where the last sublayer is a pooling layer. Our implementation is based on Torch7 \cite{TorchNIPS2011}.

\subsection{Stochastic Sampling Strategy and ``Fracking''}

Our goal is to optimize the network parameters from an arbitrarily large set of training patches. Let us consider a dataset with $N$ patches and $M\leq N$ unique 3D patch indices, each with $n_i$ associated image patches. Then, the number of matching image patches or positives $N_P$ and the number of non-matching images patches or negatives $N_N$ in the dataset is
\begin{equation}
N_P = \sum_{i=1}^M \frac{n_i (n_i-1)}{2} \qquad \text{and} \qquad N_N = \sum_{i=1}^M n_i (N-n_i)
\end{equation}
In general both $N_P$ and $N_N$ are intractable to exhaustively iterate over. We approach the problem with random sampling. For gathering positives samples we can randomly choose a set of $B_P$ 3D point indices $\{p_1,\cdots,p_{B_P}\}$, and choose two patches with corresponding 3D point indices randomly. For negatives it is sufficient to choose $B_N$ random pairs with non-matching indices.

However, when the pool of negative samples is very large random sampling will produce many negatives with a very small loss, which do not contribute to the global loss, and thus stifle the learning process. Instead, we can iterate over non-corresponding patch pairs to search for ``hard'' negatives, i.e. with a high loss. In this manner it becomes feasible to train discriminative models faster while also increasing performance. This technique is commonplace in sliding-window classification.

Therefore, at each epoch we generate a set of $B_N$ randomly chosen patch pairs, and after forward-propagation through the network and computing their loss we keep only a subset of the $B_N^M$ ``hardest'' negatives, which are back-propagated through the network in order to update the weights. Additionally, the same procedure can be used over the positive samples, i.e. we can sample $B_P$ corresponding patch pairs and prune them down to the $B_P^M$ ``hardest'' positives. We show that the combination of aggressively mining positive and negative patch pairs, which we call ``fracking'', allows us to greatly improve the discriminative capability of learned descriptors. Note that extracting descriptors with the learned models does not further require the siamese network and does not incur the computational costs related to mining.

\vspace{-1mm}
\subsection{Learning}

We normalize the dataset by subtraction of the mean of the training patches and division by their standard deviation. We then learn the weights by performing stochastic gradient descent. We use a learning rate that decreases by an order of magnitude every fixed number of iterations. Additionally, we use standard momentum in order to accelerate the learning process.  We use a subset of the data for validation, and stop training when the metric we use to evaluate the learned models converges.
Due to the exponentially large pool of positives and negatives available for training and the small number of parameters of the architectures, no techniques to cope with overfitting are used. The particulars of the learning procedure are detailed in the following section.


\vspace{-2mm}
\section{Results}
\label{sec:results}

For evaluation we use the Multi-view Stereo Correspondence dataset~\cite{BrownPAMI2011}, which consists of 64$\times$64 grayscale image patches sampled from 3D reconstructions of the Statue of Liberty (LY), Notre Dame (ND) and Half Dome in Yosemite (YO). Patches are extracted using the Difference of Gaussians detector~\cite{LoweIJCV04}, and determined as a valid correspondence if they are within 5 pixels in position, 0.25 octaves in scale and $\pi/8$ radians in angle. Figure~\ref{fig:brown-samples} shows some samples from each set, which contain significant changes in position, rotation and illumination conditions, and often exhibit very noticeable perspective changes.

\renewcommand{\imw}{1.05cm}
\begin{figure}[t]
\centering
\def\arraystretch{0}
\setlength{\tabcolsep}{0pt}
\begin{tabular}{cccccccccccc}
	\includegraphics[width=\imw]{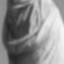}\hspace{0.5mm} &
	\includegraphics[width=\imw]{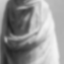}\hspace{2mm} &
	\includegraphics[width=\imw]{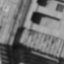}\hspace{0.5mm} &
	\includegraphics[width=\imw]{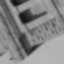}\hspace{2mm} &
	\includegraphics[width=\imw]{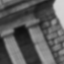}\hspace{0.5mm} &
	\includegraphics[width=\imw]{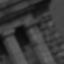}\hspace{2mm} &
	\includegraphics[width=\imw]{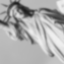}\hspace{0.5mm} &
	\includegraphics[width=\imw]{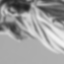}\hspace{2mm} &
	\includegraphics[width=\imw]{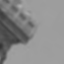}\hspace{0.5mm} &
	\includegraphics[width=\imw]{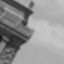}\hspace{2mm} &
	\includegraphics[width=\imw]{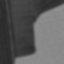}\hspace{0.5mm} &
	\includegraphics[width=\imw]{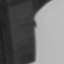}\hspace{2mm} \\
	\multicolumn{12}{c}{\vspace{1mm}} \\
	\includegraphics[width=\imw]{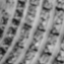}\hspace{0.5mm} &
	\includegraphics[width=\imw]{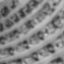}\hspace{2mm} &
	\includegraphics[width=\imw]{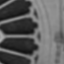}\hspace{0.5mm} &
	\includegraphics[width=\imw]{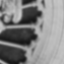}\hspace{2mm} &
	\includegraphics[width=\imw]{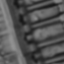}\hspace{0.5mm} &
	\includegraphics[width=\imw]{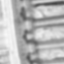}\hspace{2mm} &
	\includegraphics[width=\imw]{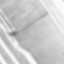}\hspace{0.5mm} &
	\includegraphics[width=\imw]{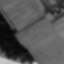}\hspace{2mm} &
	\includegraphics[width=\imw]{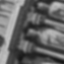}\hspace{0.5mm} &
	\includegraphics[width=\imw]{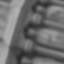}\hspace{2mm} &
	\includegraphics[width=\imw]{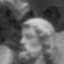}\hspace{0.5mm} &
	\includegraphics[width=\imw]{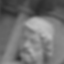}\hspace{2mm} \\
	\multicolumn{12}{c}{\vspace{1mm}} \\
	\includegraphics[width=\imw]{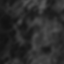}\hspace{0.5mm} &
	\includegraphics[width=\imw]{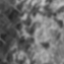}\hspace{2mm} &
	\includegraphics[width=\imw]{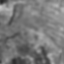}\hspace{0.5mm} &
	\includegraphics[width=\imw]{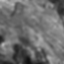}\hspace{2mm} &
	\includegraphics[width=\imw]{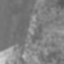}\hspace{0.5mm} &
	\includegraphics[width=\imw]{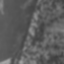}\hspace{2mm} &
	\includegraphics[width=\imw]{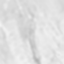}\hspace{0.5mm} &
	\includegraphics[width=\imw]{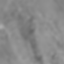}\hspace{2mm} &
	\includegraphics[width=\imw]{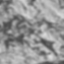}\hspace{0.5mm} &
	\includegraphics[width=\imw]{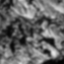}\hspace{2mm} &
	\includegraphics[width=\imw]{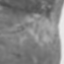}\hspace{0.5mm} &
	\includegraphics[width=\imw]{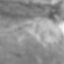}\hspace{2mm} \\
\end{tabular}
\caption{Pairs of corresponding samples from the Multi-view Stereo Correspondence dataset. Top row: `Liberty' (LY). Middle row: `Notre Dame' (ND). Bottom row: `Yosemite' (YO).}
\label{fig:brown-samples}
\vspace{-5mm}
\end{figure}

We join the data from LY and YO to form a training set with over a million patches. Out of these we reserve a subset of 10,000 unique 3D points for validation (roughly 30,000 patches). 
The resulting training dataset contains 1,133,525 possible positive patch combinations and 1.117$\times 10^{12}$ possible negative combinations. This skew is common in correspondence problems such as stereo or structure from motion; we address it with aggressive mining or ``fracking".


A popular metric for classification systems is the Receiving Operator Characteristic (ROC), used e.g. in~\cite{BrownPAMI2011}, which can be summarized by its area under the curve (AUC). However, ROC curves can be misleading when  the number of positive and negative samples are very different~\cite{DavisICML06}, and is already nearly saturated for the baseline descriptor, SIFT (see Sec.~\ref{sec:appendix1}). A richer metric is the Precision-Recall curve (PR). We benchmark our models with PR curves and their AUC. In particular, for each of the 10,000 unique points in the validation set we randomly sample two corresponding patches and 1,000 non-corresponding patches, and use them to compute the PR curve. We rely on the validation set for the LY+YO split to examine different configurations, network architectures and mining techniques; these results are presented in Secs.~\ref{sec:results:depth}-\ref{sec:results:filter-dims}.

\begin{table}[t!]
\parbox{.45\linewidth}{
	\centering
	\renewcommand{\arraystretch}{1.2}
	\begin{tabular}{lcc}
	\toprule
	Architecture & Parameters & PR AUC \\
	\midrule
	SIFT & --- & 0.361 \\
	CNN1\_NN1 & 68,352 & 0.032 \\
	CNN2 & 27,776 & 0.379 \\
	CNN2a\_NN1 & 145,088 & 0.370 \\
	CNN2b\_NN1 & 48,576 & 0.439 \\
	CNN3\_NN1 & 62,784 & 0.289 \\
	CNN3 & 46,272 & {\bf 0.558} \\
	\bottomrule
	\end{tabular}
	\caption{Effect of network depth, and fully convolutional networks vs networks with a fully-connected layer. PR AUC on the validation set for the top-performing iteration.}
	\label{tbl:depth}
}
\hfill
\parbox{.5\linewidth}{
	\centering
	\renewcommand{\arraystretch}{1.2}
	\begin{tabular}{lc}
	\toprule
	Architecture & PR AUC \\
	\midrule
	SIFT & 0.361 \\
	CNN3 & {\bf 0.558} \\
	CNN3 ReLU & 0.442 \\
	CNN3 No Norm & 0.511 \\
	CNN3 MaxPool & 0.420 \\
	\bottomrule
	\end{tabular}
	\caption{Fully convolutional CNN3 models with Tanh and ReLU, without normalization, and with max pooling instead of L$_2$ pooling. The best results are obtained for Tanh units, normalization, and L$_2$ pooling (i.e. `CNN3'). PR AUC on the validation set for the top-performing iteration.}
	\label{tbl:units}
}
\end{table}

\begin{figure}[t!]
\begin{minipage}[t]{0.48\linewidth}
	\includegraphics[width=6.3cm,height=4.9cm]{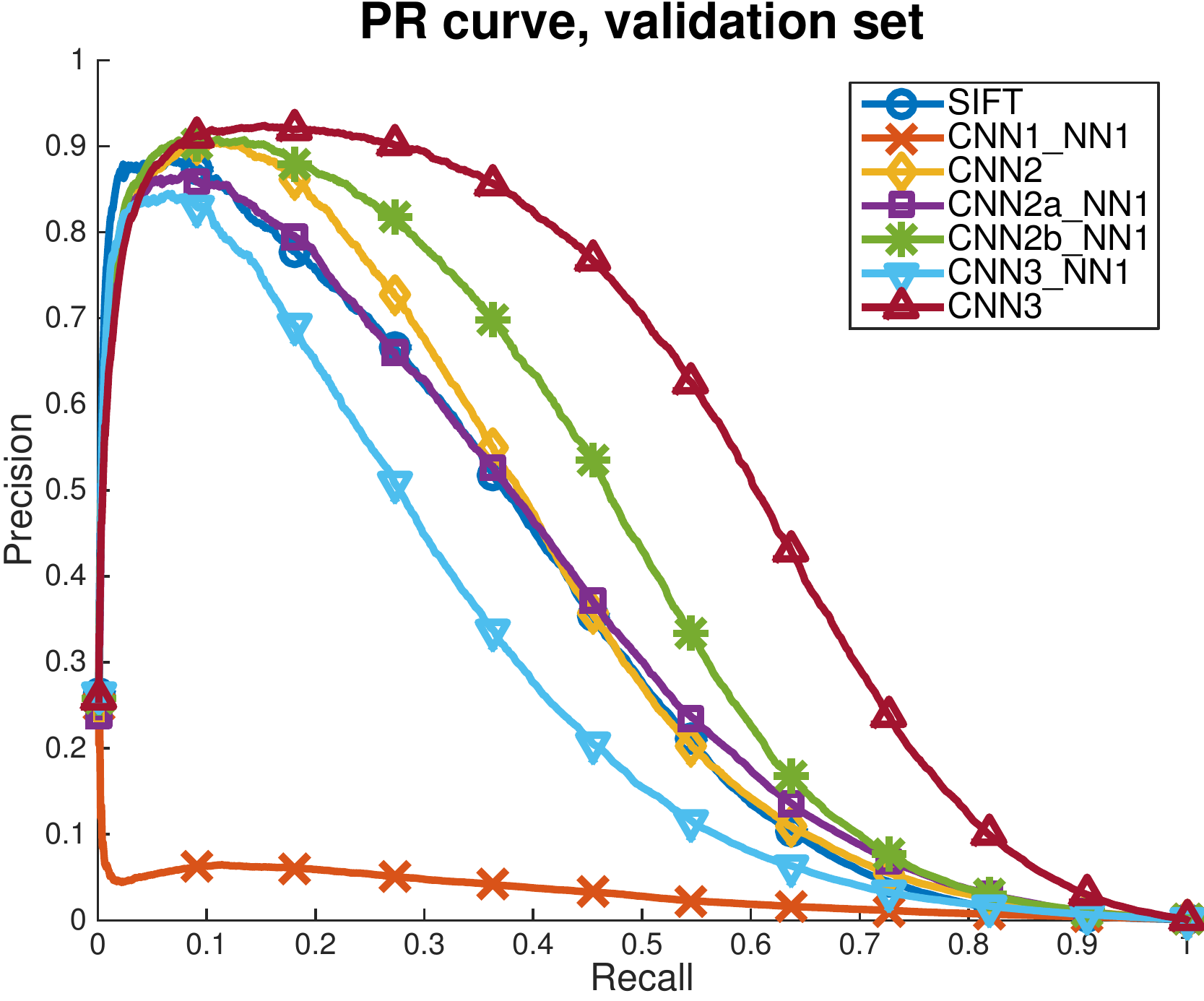}
	\caption{Effect of network depth (up to 3 CNN layers), and fully convolutional networks vs networks with final fully-connected layer (NN1). Fully-convolutional models outperform models with fully-connected neural network at the end.}
	\label{fig:depth}
\end{minipage}
\hfill
\begin{minipage}[t]{0.48\linewidth}
	\includegraphics[width=6.3cm,height=4.9cm]{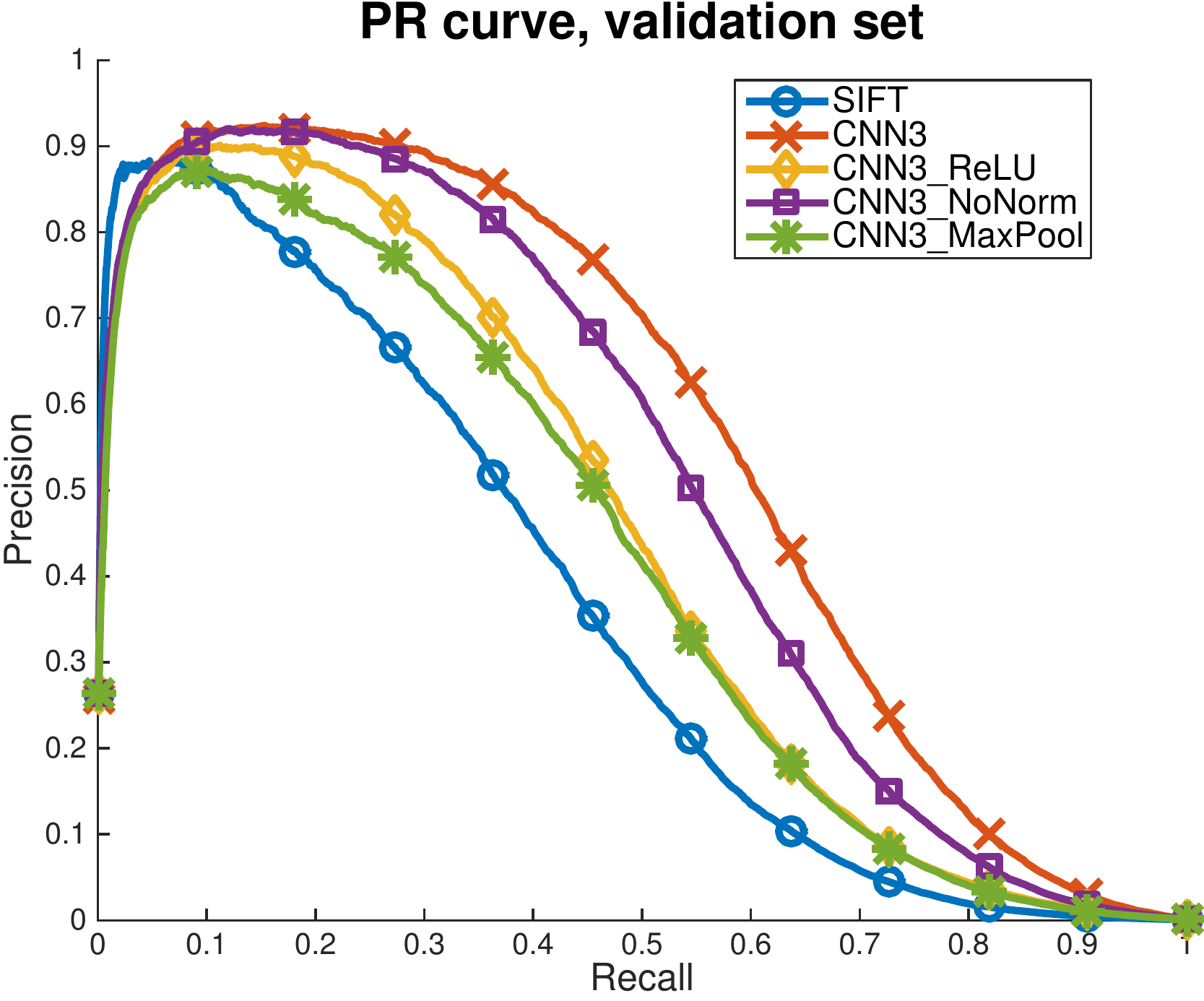}
	\caption{Fully convolutional CNN3 models with Tanh and ReLU, without normalization, and with max pooling instead of L$_2$ pooling. The best results are obtained for Tanh units, normalization, and L$_2$ pooling (`CNN3' model).}
	\label{fig:units}
\end{minipage}%
\vspace{-5mm}
\end{figure}

Finally, we evaluate the top-performing models over the test set in Sec.~\ref{sec:results:generalization}. We follow the same procedure as for validation, compiling the results for 10,000 points with 1,000 non-corresponding matches each, now over 10 different folds (see Sec.~\ref{sec:appendix1} for details). We run three different splits, for generalization: LY+YO (tested on ND), LY+ND (tested on YO), and YO+ND (tested on LY).

We will consider all hyperparameters to be the same unless otherwise mentioned, i.e. a learning rate of $0.01$ that decreases ever $10,000$ iterations by a factor of 10. We consider negative mining with $B_N=256$ and $B_N^M=128$, and no positive mining; i.e. $B_P=B_P^M=128$.

\vspace{-2mm}
\subsection{Depth and Fully Convolutional Architectures}
\label{sec:results:depth}

The network depth is constrained by the size of the patch. We consider only up to 3 convolutional layers (CNN\{1-3\}). Additionally, we consider adding a single fully-connected layer at the end (NN1). Fully-connected layers increase the number of parameters by a large factor, which increases the difficulty of learning and can lead to overfitting. We show the results of the various architectures we evaluate in Table~\ref{tbl:depth} and Fig.~\ref{fig:depth}. Deeper networks outperform shallower ones, and architectures with a fully-connected layer at the end do worse than fully convolutional architectures. In the following experiments with consider only models with 3 convolutional layers.

\begin{table}
\parbox{.53\linewidth}{
	\small
	\centering
	\renewcommand{\arraystretch}{0.9}
	\setlength{\tabcolsep}{8pt}
	\begin{tabular}{cccccc}
	\toprule
	$B_N$ & $B_P$ & $R_P$ & $R_N$ & Cost & PR AUC \\
	\midrule
	128 & 128 & 1 & 1 & --- & 0.366 \\
	256 & 256 & 1 & 1 & --- & 0.374 \\
	512 & 512 & 1 & 1 & --- & 0.369 \\
	1024 & 1024 & 1 & 1 & --- & 0.325 \\
	\midrule
	128 & 256 & 1 & 2 & 20\% & 0.558 \\
	256 & 256 & 2 & 2 & 35\% & 0.596 \\
	512 & 512 & 4 & 4 & 48\% & 0.703 \\
	1024 & 1024 & 8 & 8 & 67\% & {\bf 0.746} \\
	2048 & 2048 & 16 & 16 & 80\% & 0.538 \\
	\bottomrule
	\end{tabular}
	\caption{Four top rows: increasing batch size without mining. Four bottom rows: mining. Mining factors indicate the samples considered ($B_P$, $B_N$): 128 are used for training. Column 5 indicates the fraction of the computational cost spent mining hard samples.}
	\label{tbl:fracking}
}
\hfill
\parbox{.42\linewidth}{
	\small
	\centering
	\renewcommand{\arraystretch}{1.5}
	\setlength{\tabcolsep}{3pt}
	\begin{tabular}{lccc}
	\toprule
	Architecture & Output & Parameters & PR AUC \\
	\midrule
	SIFT & 128f & --- & 0.361 \\
	CNN3 & 128f & 46,272 & {\bf 0.596} \\
	CNN3 Wide & 128f & 110,496 & 0.552 \\
	CNN3\_NN1 & 128f & 62,784 & 0.456 \\
	CNN3\_NN1 & 32f & 50,400 & 0.389 \\
	\bottomrule
	\end{tabular}
	\caption{Effect of the size of the filters in the CNN3 model, and the fully-connected layers at the end. The best results are obtained with fully-convolutional networks.}
	\label{tbl:filter-dims}
}
\end{table}

\begin{figure}[t!]
\begin{minipage}[c]{0.32\linewidth}
	\includegraphics[width=4.0cm]{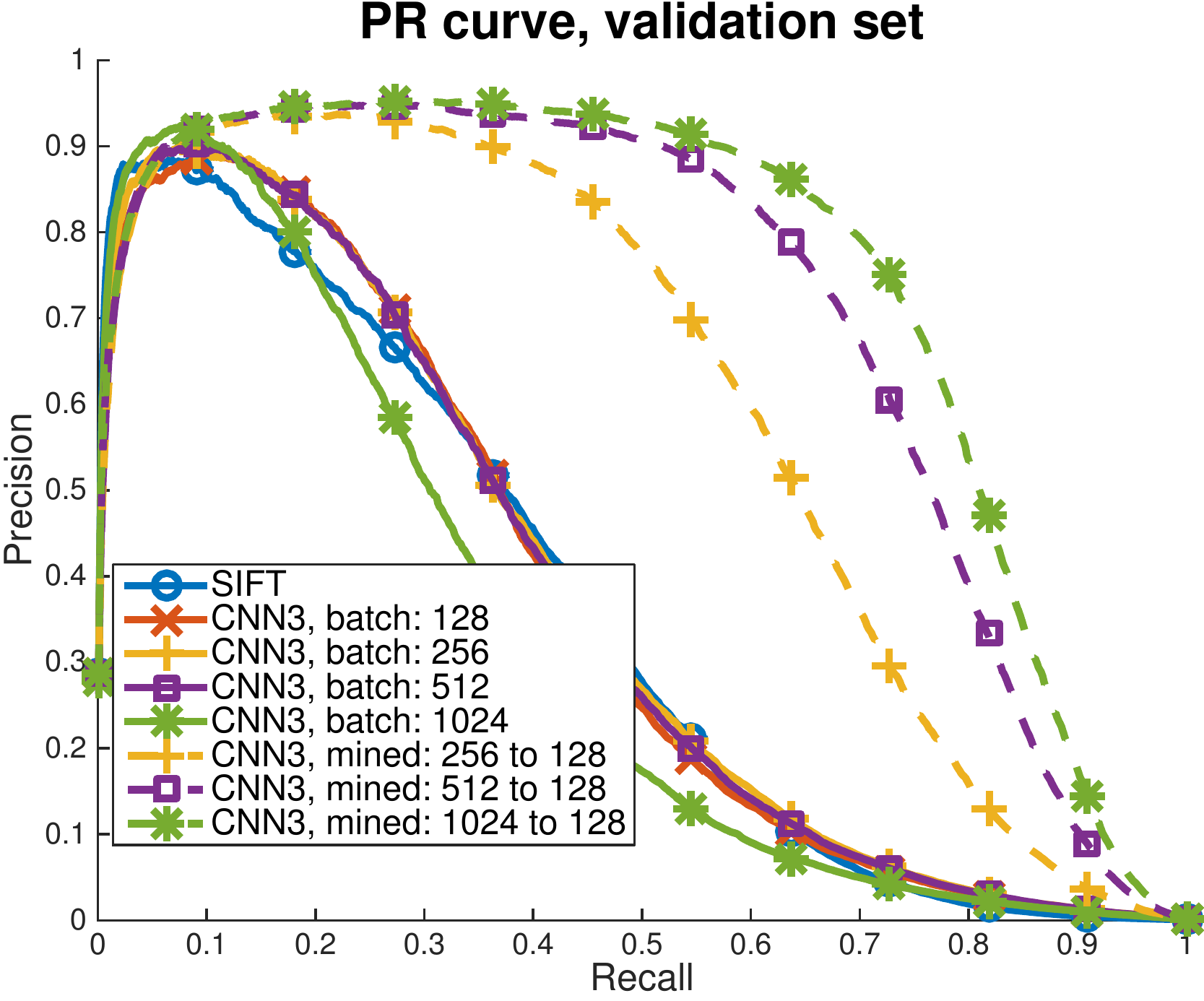}
	\caption{`Fracking' vs comparable batch size. Notice that larger pools are not helpful.}
	\vspace{-4mm}
	\label{fig:fracking-vs-batchsize}
\end{minipage}
\hfill
\begin{minipage}[c]{0.32\linewidth}
	\includegraphics[width=4.0cm]{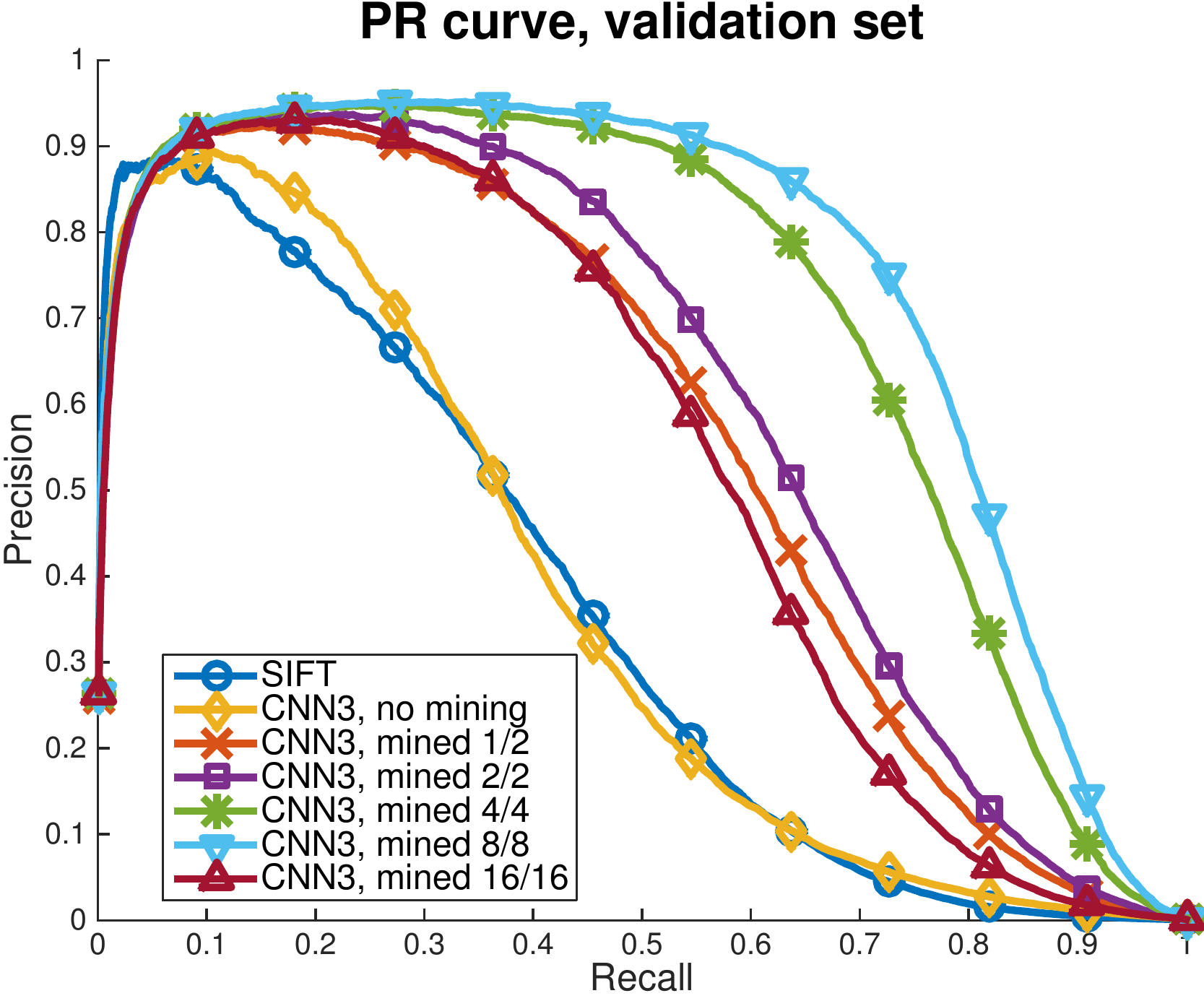}
	\caption{Effect of mining. The best results are obtained with 8/8 factors (from 1024 to 128).}
	\vspace{-4mm}
	\label{fig:fracking}
\end{minipage}
\hfill
\begin{minipage}[c]{0.32\linewidth}
	\includegraphics[width=4.0cm]{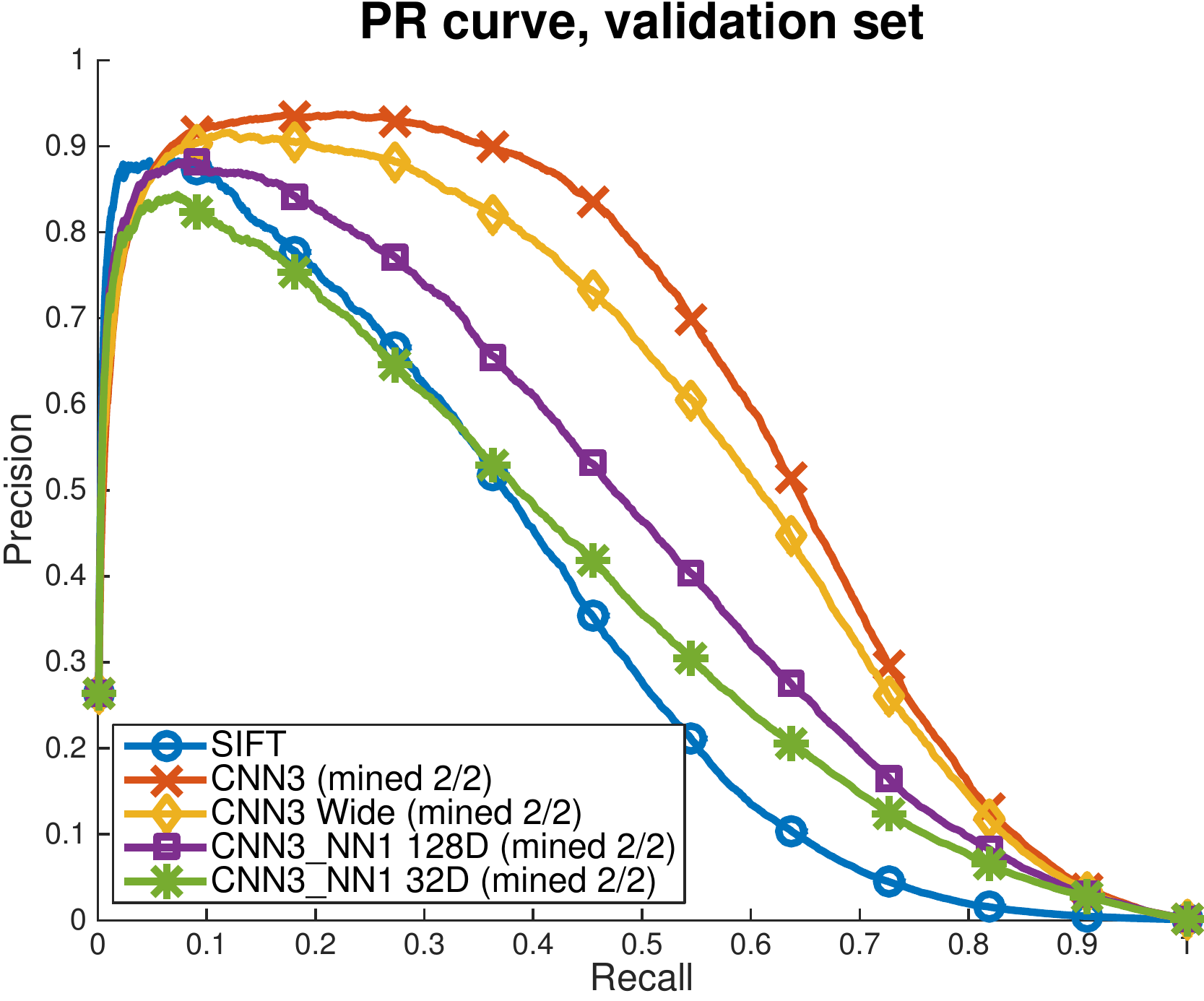}
	\caption{Filter size and fully-connected layers. Fully-convolutional networks do best.}
	\vspace{-4mm}
	\label{fig:filter-dims}
\end{minipage}
\end{figure}

\vspace{-2mm}
\subsection{Hidden Units   Mapping, Normalization, and Pooling}
\label{sec:results:units}

It is generally accepted that Rectified Linear Units (ReLU) perform much better in classification tasks~\cite{KrizhevskyNIPS12} than other non-linear functions. They are, however, ill-suited for tasks with continuous outputs such as regression tasks or the problem at hand, as they can only output positive values. We consider both the standard Tanh and ReLU. For the ReLU case we still use Tanh for the last layer. We also consider not using the normalization sublayer for each of the convolutional layers. Finally, we consider using max pooling rather than L$_2$ pooling. We show results for the fully-convolutional CNN3 architecture in Table~\ref{tbl:units} and Fig.~\ref{fig:units}. The best results are obtained with Tanh, normalization and L$_2$ pooling (`CNN3' in the table/plot). We will use this configuration in the following experiments.

\vspace{-2mm}
\subsection{Fracking}
\label{sec:results:fracking}

We analyze the effect of both positive and negative mining, or ``fracking'', by training different models in which a large, initial pool of $B_P$ positives and  $B_N$ negatives are pruned to a smaller number of `hard' positive and negative matches, used to update the parameters of the network. We observe that increasing the batch size does not offer benefits in training: see Table~\ref{tbl:fracking} and Fig~\ref{fig:fracking-vs-batchsize}. We thus keep the batch size fixed to $B_N^M=128$ and $B_P^M=128$, and increase the ratio of both negative mining $R_N = B_N/B_N^M$ and positive mining $R_P = B_P/B_P^M$. We keep all other parameters constant. We use the notation $R_P$/$R_N$, for brevity.

Large mining factors have a high computational cost, up to 80\% of the total computational cost, which includes mining  (i.e. forward propagation of all $B_P$ and $B_N$ samples)  and learning (i.e. backpropagating the ``hard'' positive and negative samples). In order to speed up the learning process we initialize the CNN3 models with positive mining, i.e. 2/2, 4/4, 8/8 and 16/16, with an early iteration of a model trained only with negative mining (1/2).

Results are shown in Table~\ref{tbl:fracking} and Fig.~\ref{fig:fracking}. We see that for this particular problem aggressive mining is fundamental. This is likely due to the extremely large number of both negatives and positives in the dataset, in combination with models with a low number of parameters. We observe a drastic increase in performance up to 8/8 mining factors.

\begin{table}[t!]
\centering
\begin{tabular}{llcccccc}
	\toprule
	\multirow{2}{*}{Train} & \multirow{2}{*}{Test} & \multirow{2}{*}{SIFT} & CNN3 & CNN3 & CNN3 & CNN3 & PR AUC Increase\\
	& & & mine-1/2 & mine-2/2 & mine-4/4 & mine-8/8 & (Best vs SIFT)\\
	\midrule
	LY+YOS & ND & 0.349 & 0.535 & 0.555 & 0.630 & {\bf 0.667} & 91.1\% \\
	LY+ND & YOS & 0.425 & 0.383 & 0.390 & 0.502 & {\bf 0.545} & 28.2\% \\
	YOS+ND & LY & 0.226 & 0.460 & 0.483 & 0.564 & {\bf 0.608} & 169.0\% \\
	\bottomrule
\end{tabular}
\caption{PR AUC for the generalized results over the three dataset splits, for different mining factors.}
\label{tbl:splits}
\vspace{-1mm}
\end{table}

\begin{figure}[t!]
\centering
\includegraphics[width=0.32\linewidth]{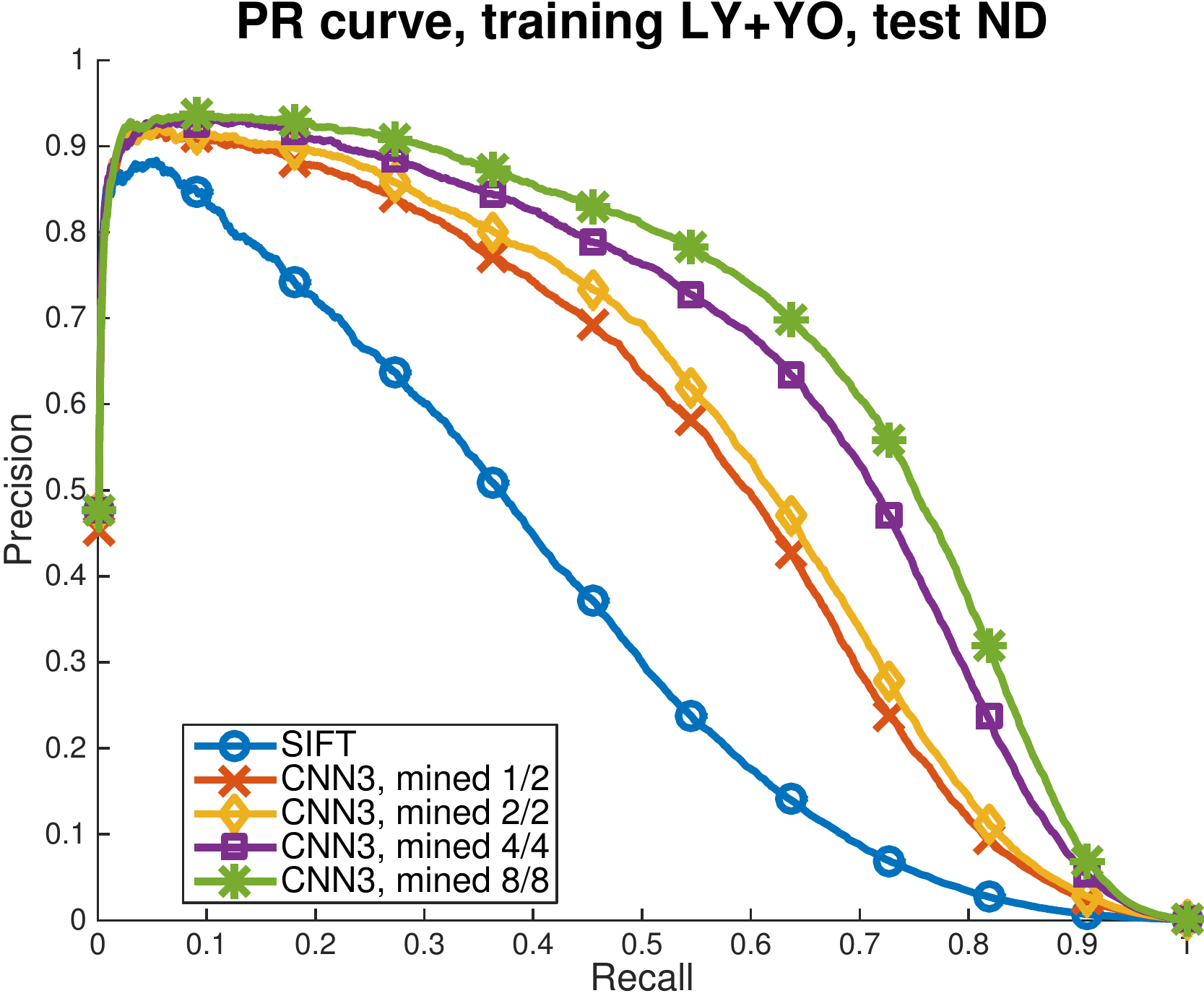}
\includegraphics[width=0.32\linewidth]{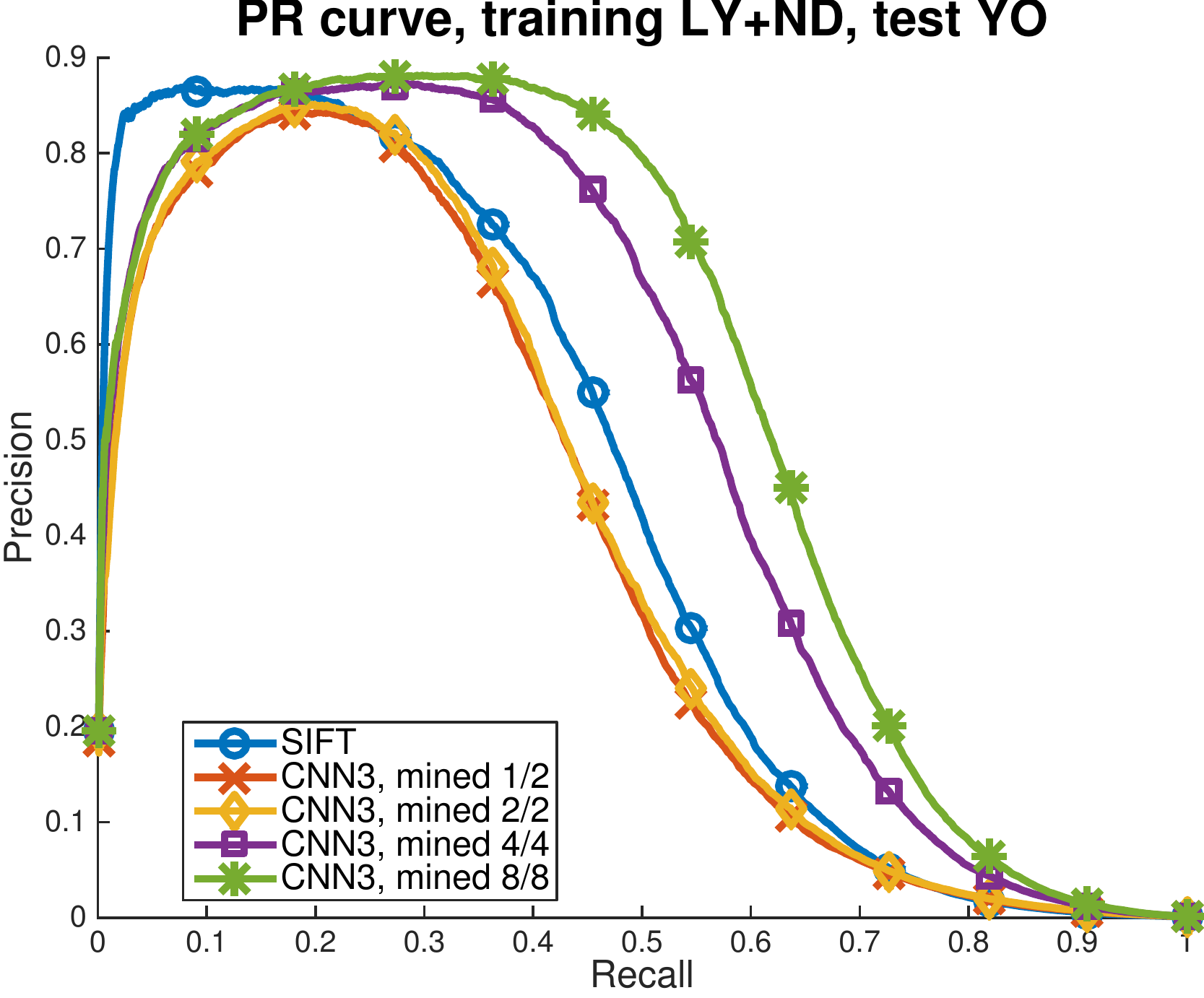}
\includegraphics[width=0.32\linewidth]{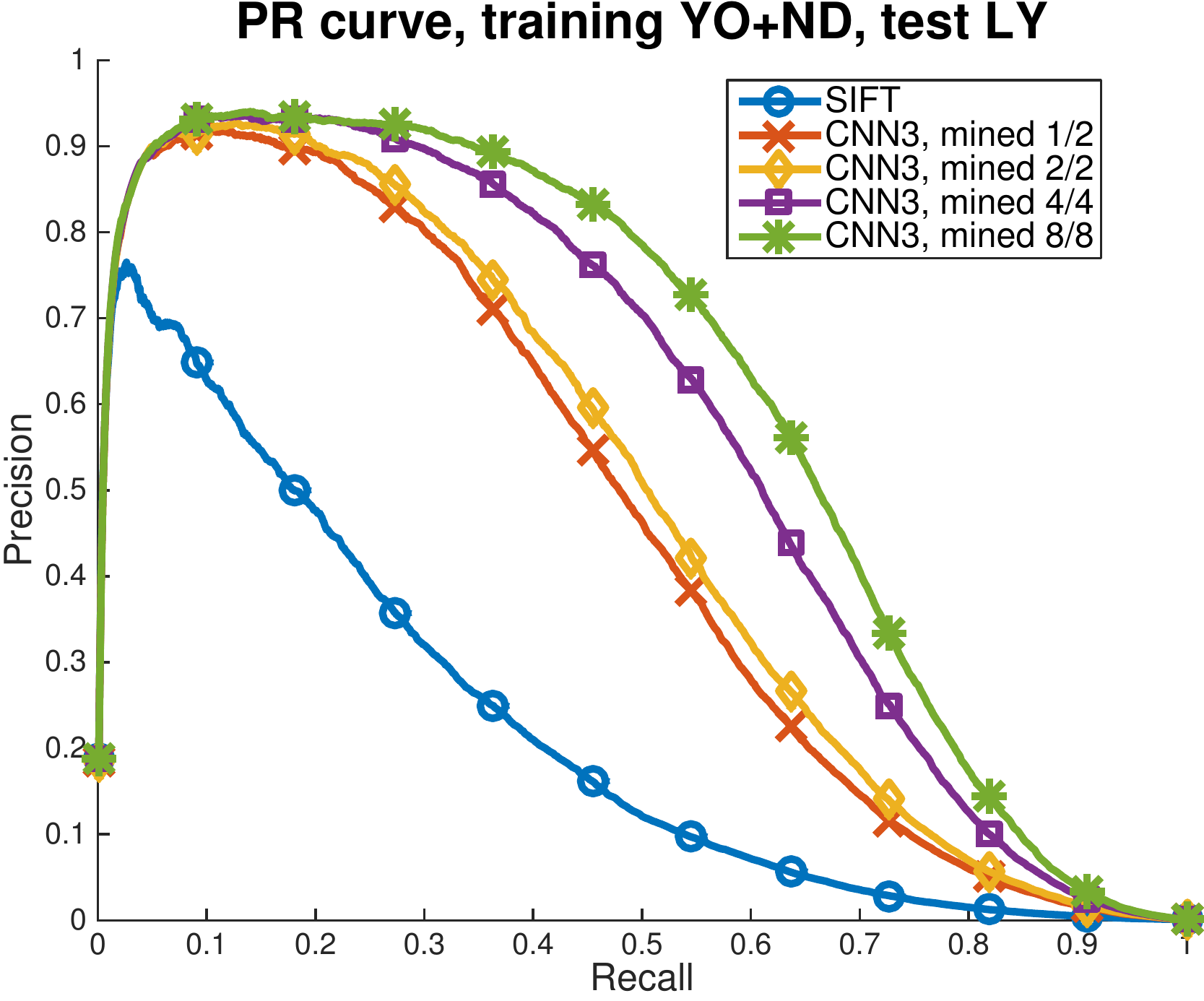}
\caption{PR curves for the generalized results over the three dataset splits.}
\label{fig:splits}
\vspace{-2mm}
\end{figure}

\begin{table}[t!]
\centering
\begin{tabular}{lcccccccc}
	\toprule
	\multirow{2}{*}{Test} & SIFT & BGM & L-BGM & \multicolumn{3}{c}{BinBoost-\{64,128,256\}} & Ours (best) & PR AUC Increase \\
	& (128f) & (256b) & (64f) & (64b) & (128b) & (256b) & (128f) & (Ours vs Best) \\
	\midrule
	ND  & 0.349 & 0.487 & 0.495 & 0.267 & 0.451 & 0.549 & {\bf 0.667} & 21.5\%\\
	YOS & 0.425 & 0.495 & 0.517 & 0.283 & 0.457 & 0.533 & {\bf 0.545} &  0.2\%\\
	LY  & 0.226 & 0.268 & 0.355 & 0.202 & 0.346 & 0.410 & {\bf 0.608} & 48.3\%\\
	\bottomrule
\end{tabular}
\caption{PR AUC for the generalized results over the three dataset splits, compared to \cite{TrzcinskiCVPR2013}.}
\label{tbl:splits-binboost}
\vspace{-1mm}
\end{table}

\begin{figure}[t!]
\centering
\includegraphics[width=0.32\linewidth]{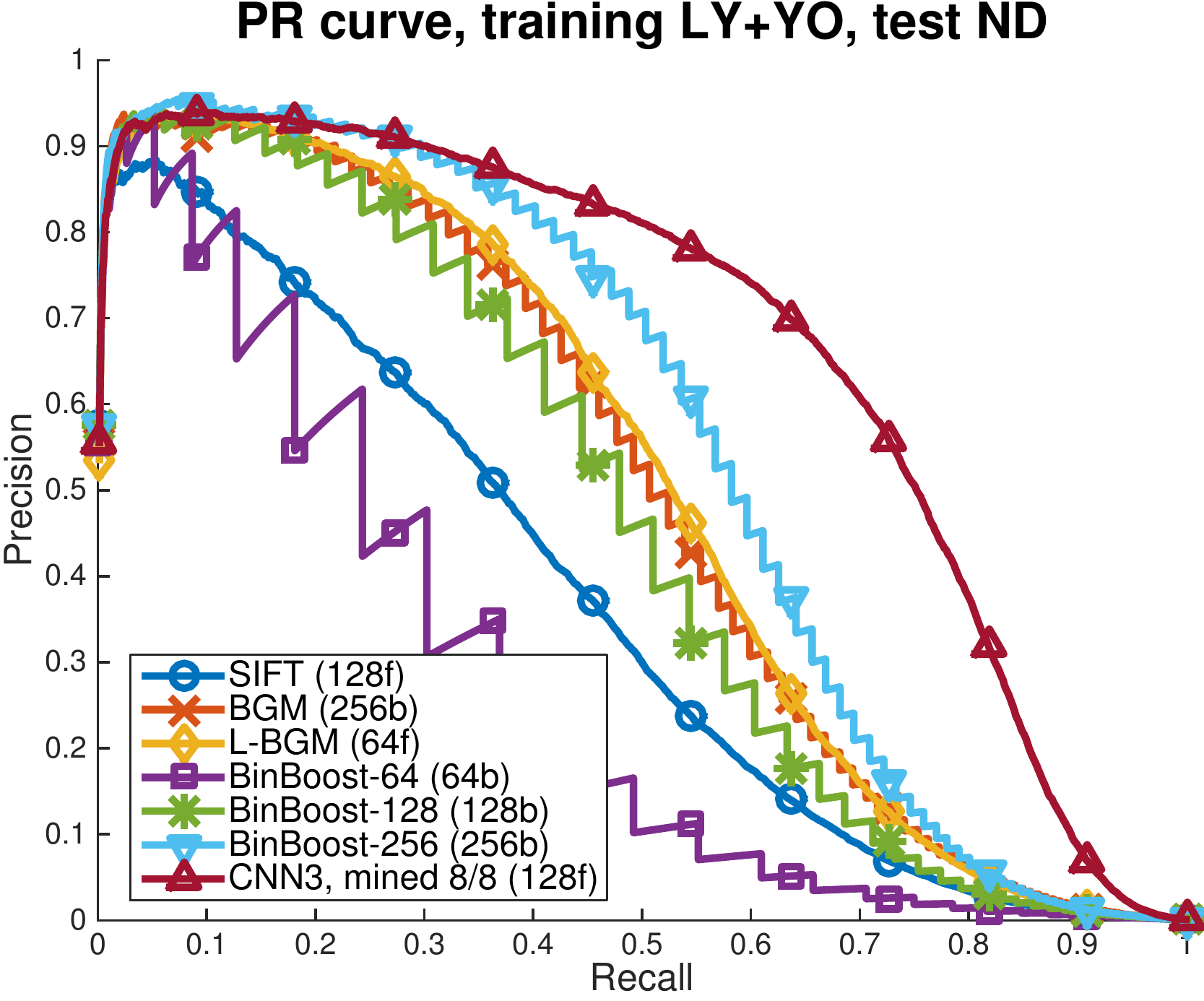}
\includegraphics[width=0.32\linewidth]{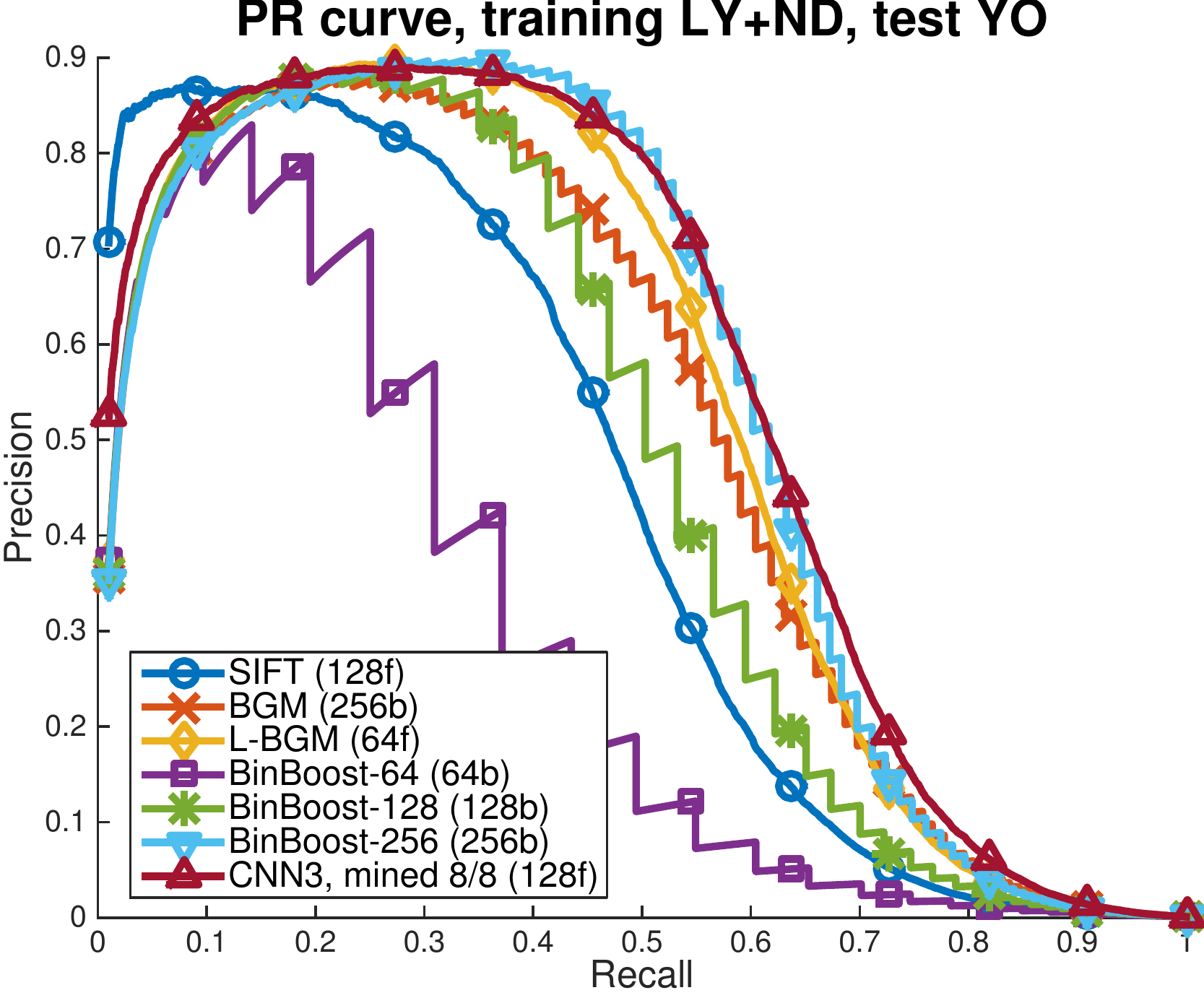}
\includegraphics[width=0.32\linewidth]{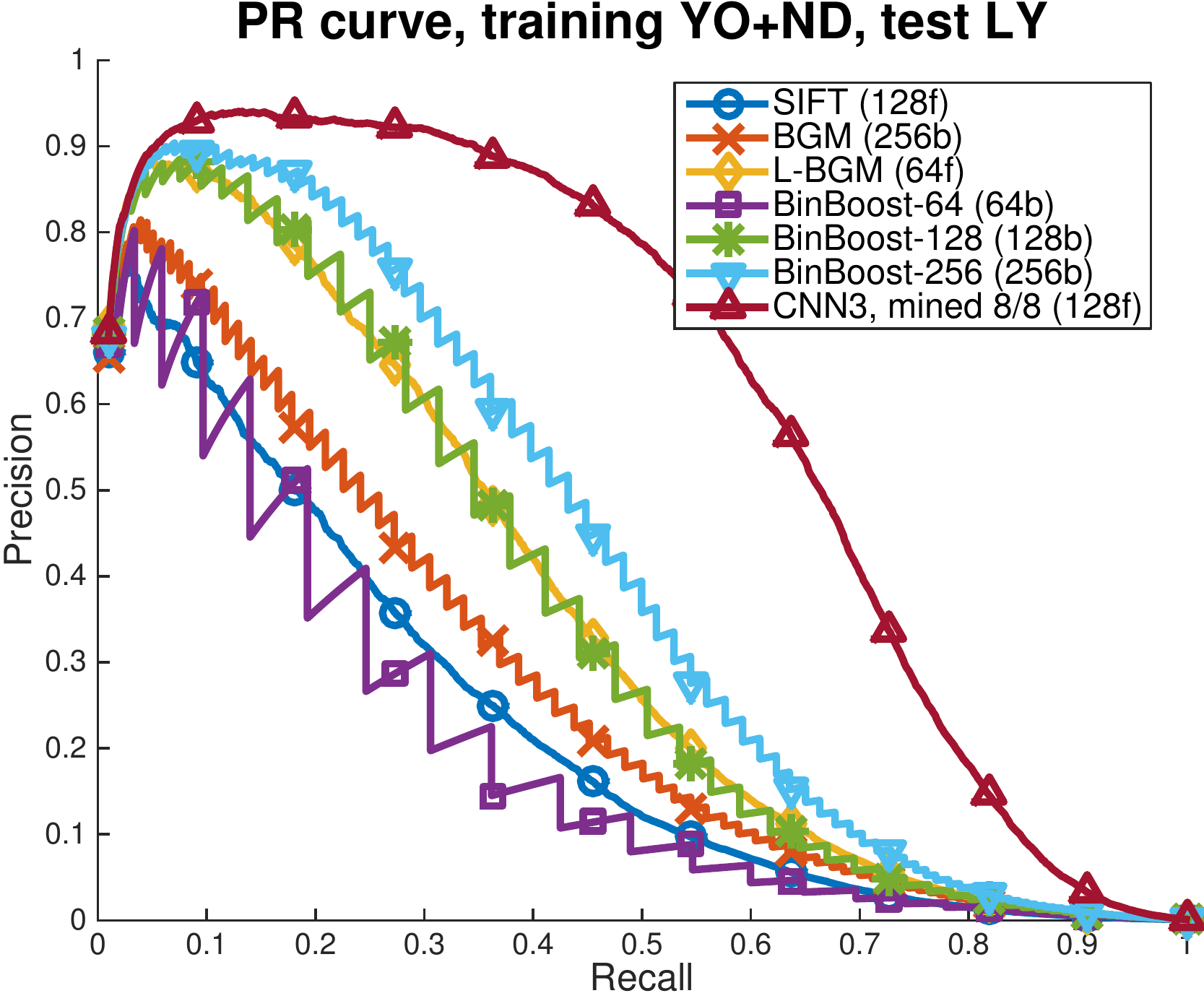}
\caption{PR curves for the generalized results over the three dataset splits, compared to \cite{TrzcinskiCVPR2013}.}
\label{fig:splits-binboost}
\vspace{-4mm}
\end{figure}

\vspace{-2mm}
\subsection{Number of Filters and Descriptor Dimension}
\label{sec:results:filter-dims}

We analyze increasing the number of filters in the CNN3 model, and adding a fully-connected layer that can be used to decrease the dimensionality of the descriptor. We consider increasing the number of filters in layers 1 and 2 from 32 and 64 to 64 and 96, respectively. Additionally, we double the number of internal connections between layers. This more than doubles the number of parameters in this network. To analyze descriptor dimensions we consider the CNN3\_NN1 model and change the number of outputs in the last fully-connected layer from 128 to 32. In this case we consider positive mining with $B_P=256$ (i.e. 2/2). Results can be seen in Table~\ref{tbl:filter-dims} and Fig.~\ref{fig:filter-dims}. The best results are obtained with smaller filters and fully-convolutional networks. Additionally we notice that mining is also instrumental for models the NN1 layer (compare results with Table~\ref{tbl:depth}).

\vspace{-1mm}
\subsection{Generalization and Comparison to State-of-the-Art}
\label{sec:results:generalization}

In this section we consider the three dataset splits. We train the best performing models, i.e. CNN3 with different mining ratios, on a combination of two sets, and test them on the remaining set. We select the training iteration that performs best over the validation set. The test datasets are very large (up to 633K patches) and we use the same procedure as for validation, evaluating 10,000 unique points, each with 1,000 random non-corresponding matches. We repeat this process over 10 folds, thus considering 100,000 sets of one corresponding patch vs 1,000 non-corresponding patches. We show results in terms of PR AUC in Table~\ref{tbl:splits}, and the corresponding PR curves are pictured in Fig.~\ref{fig:splits}.

We report consistent improvements over the baseline, i.e. SIFT. The performance varies significantly from split to split; this is due to the nature of the different sets. `Yosemite' contains mostly frontoparallel translations with illumination changes and no occlusions (Fig.~\ref{fig:brown-samples}, row 3); SIFT performs well on this type of data. Our learned descriptors outperform SIFT on the high-recall regime (over 20\% of the samples; see Fig.~\ref{fig:splits}), and is 28\% better overall in terms of PR AUC. The effect is much more dramatic on `Notredame' and `Liberty', which contains significant patch translation and rotation, as well as viewpoint changes around outcropping  non-convex objects which result in occlusions (see Fig.~\ref{fig:brown-samples}, rows 1-2). Our learned descriptors outperform SIFT by 91\% and 169\% testing over ND and LY, respectively.

We additionally compare against the state-of-the-art~\cite{TrzcinskiCVPR2013}. In particular, we compare against 4 binary descriptor variants (BGM, BinBoost-64, BinBoost-128, and BinBoost-256) and L-BGM, as well as SIFT. For the binary descriptors we use the Hamming distance instead of Euclidean distance. The results are summarized in Table~\ref{tbl:splits-binboost} and shown in Fig.~\ref{fig:splits-binboost}. Our approach outperforms all descriptors, with the largest relative improvement on the `Liberty' (LY) dataset. Due to the binary nature of the Hamming distance, the curves for the binary descriptors can be seen to have a sawtooth shape where each tooth corresponds to a 1-bit difference.

\vspace{-1mm}
\subsection{Qualitative analysis}
\label{sec:results:qualitative}

\begin{figure}
\centering
\includegraphics[height=5.8cm,width=6cm]{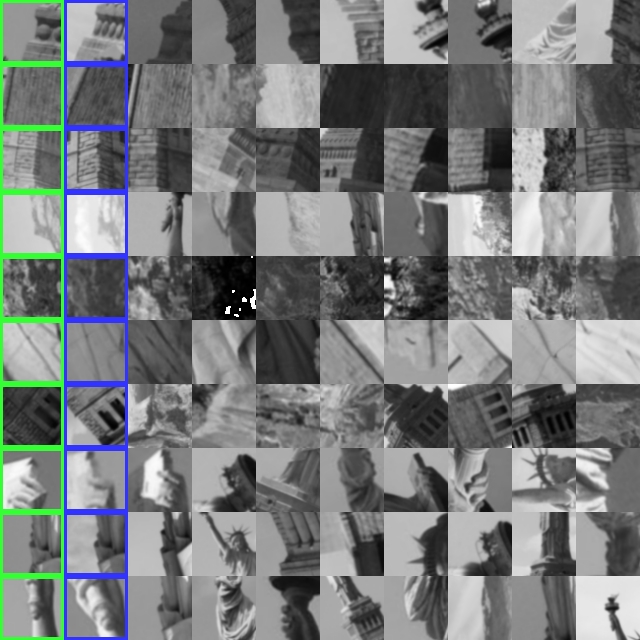} \hspace{6mm}
\includegraphics[height=5.8cm,width=6cm]{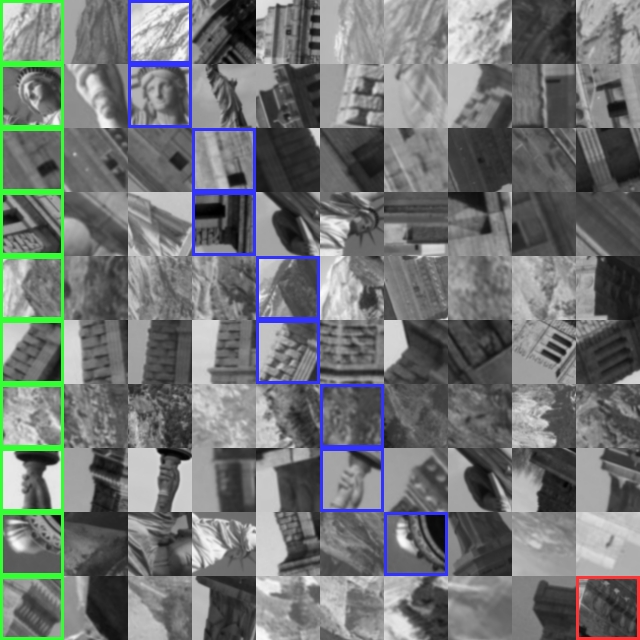}
\caption{Samples of matches retrieved with our descriptor. Each row depicts the reference patch (in green) and the top matching candidates, sorted from left to right by decreasing similarity. The ground truth match is highlighted in blue. {\bf Left:} Examples where the ground truth match is retrieved in the first position. {\bf Right:} Examples in which the ground truth match is ranked at positions 2-6. The last row shows a failure case, highlighted in red, where the correct match is ranked 632/1000.}
\label{fig:ranked-matches}
\vspace{-5mm}
\end{figure}

Figure~\ref{fig:ranked-matches} shows samples of matches retrieved with our CNN3-mined-4/4 network, over the validation set for the first split. In this experiment the corresponding patches were ranked in the first position in 76.5\% of cases; a remarkable result, considering that every true match had to be chosen from a pool of 1,000 false correspondences. The right-hand image shows cases where the ground truth match was not ranked first; notice that most of these patches exhibit significant changes of appearance. We include a failure case (highlighted in red), caused by a combination of large viewpoint and illumination changes; these misdetections are however very uncommon.

\vspace{-3mm}
\section{Conclusions}
\label{sec:conclusions}

We use siamese networks to train deep convolutional models for the extraction of image descriptors for correspondence matching. This problem typically involves small patches and large volumes of data. The former constrains the size of the network, limiting the discriminating power of the models, while the latter makes it intractable to exhaustibly iterate over all the training samples. We address this problem with a novel training scheme based on aggressive mining of both positive and negative correspondences.

Current research in convolutional neural networks for computer vision is focused on classification rather than regression. Despite previous research in this area~\cite{JahrerCVWW2008,OsendorferICONIP13} it remains unclear what is the most adequate architecture for the problem at hand. This is a critical point, as evidenced by the dramatic effect \citenop{KrizhevskyNIPS12} had in re-invigorating the field. In this paper we investigate a wide range of architectures (filter size, Tanh and ReLU units, normalization), and consider both fully-convolutional networks and networks with a fully-connected neural network at the end.

We show that the combination of state-of-the-art techniques with aggressive mining in the training stage can result in large performance gains. We obtain  up to over 2.5x the performance of SIFT, and up to 1.5x the performance of the state-of-the-art in terms of the Precision-Recall AUC. Our experiments over different data splits suggest that learning descriptors is particularly relevant for hard correspondence problems (e.g. `Liberty' and `Notredame' sets). The best models are fully convolutional.

We identify multiple directions for future research. Color has proved informative in deep learning for detection problems; our current models are built on grayscale data due to data restrictions. Our networks are likewise restricted by the size of the patches ($64\times64$ pixels), so that we currently limit our models to three convolutional layers. We intend to explore the performance of deeper networks along with larger patches. Additionally, our study indicates that fully-convolutional models outperform models with a fully-connected neural network at the end, but we intend to study this problem in further detail, particularly with dimensionality reduction in mind.

\comment{
We identify multiple directions for future research. Color has proved informative in deep learning for detection problems; our current models are built on grayscale data due to data restrictions. Our networks are likewise restricted by the size of the patches ($64\times64$ pixels), so that we currently limit our models to three convolutional layers. We intend to explore the performance of deeper networks along with larger patches; 
it is particularly interesting to investigate the effects of occlusions, which then become inevitable. Our study indicates that fully-convolutional models outperform models with a fully-connected neural network at the end, but we intend to study this problem in further detail, particularly with dimensionality reduction in mind. Finally, it remains unclear how to train these models for other applications, particularly for recognition. We have not compared our CNN-based models to other learned descriptors~\cite{SimonyanPAMI14} due to time constraints, but we intend to do so in the future.}



\vspace{-2mm}
\subsubsection*{Acknowledgments}
\vspace{-2mm}
This project was partially funded by the ERA-net CHISTERA project VISEN (PCIN-2013-047), ARCAS (FP7-ICT-2011-287617),  PAU+ (DPI2011-27510), grant ANR-10-JCJC-0205 (HICORE), MOBOT (FP7-ICT-2011-600796) and RECONFIG (FP7-ICT-600825). We gratefully acknowledge the support of NVIDIA Corporation with the donation of the GPUs used for this research.

\vspace{-4mm}
\setlength{\bibsep}{2pt}
\bibliographystyle{iclr2015}
\bibliography{top}

\begin{thebibliography}{24}
\providecommand{\natexlab}[1]{#1}
\providecommand{\url}[1]{\texttt{#1}}
\expandafter\ifx\csname urlstyle\endcsname\relax
  \providecommand{\doi}[1]{doi: #1}\else
  \providecommand{\doi}{doi: \begingroup \urlstyle{rm}\Url}\fi

\bibitem[Bay et~al.(2006)Bay, Tuytelaars, and {Van Gool}]{BayECCV2006}
Bay, H., Tuytelaars, T., and {Van Gool}, L.
\newblock {SURF: Speeded Up Robust Features}.
\newblock In \emph{ECCV}, 2006.

\bibitem[Bromley et~al.(1994)Bromley, Guyon, Lecun, Säckinger, and
  Shah]{BromleyNIPS94}
Bromley, J., Guyon, I., Lecun, Y., Säckinger, E., and Shah, R.
\newblock Signature verification using a "siamese" time delay neural network.
\newblock In \emph{NIPS}, 1994.

\bibitem[Brown et~al.(2011)Brown, Hua, and Winder]{BrownPAMI2011}
Brown, M., Hua, Gang, and Winder, S.
\newblock Discriminative learning of local image descriptors.
\newblock \emph{PAMI}, 33\penalty0 (1):\penalty0 43--57, 2011.

\bibitem[Collobert et~al.(2011)Collobert, Kavukcuoglu, and
  Farabet]{TorchNIPS2011}
Collobert, R., Kavukcuoglu, K., and Farabet, C.
\newblock Torch7: A matlab-like environment for machine learning.
\newblock In \emph{BigLearn, NIPS Workshop}, 2011.

\bibitem[Culurciello et~al.(2013)Culurciello, Jin, Dundar, and
  Bates]{CulurcielloARXIV13}
Culurciello, E., Jin, J., Dundar, A., and Bates, J.
\newblock An analysis of the connections between layers of deep neural
  networks.
\newblock \emph{CoRR}, abs/1306.0152, 2013.

\bibitem[Davis \& Goadrich(2006)Davis and Goadrich]{DavisICML06}
Davis, J. and Goadrich, M.
\newblock The relationship between {PR} and {ROC} curves.
\newblock In \emph{ICML}, 2006.

\bibitem[Felzenszwalb et~al.(2010)Felzenszwalb, Girshick, McAllester, and
  Ramanan]{FelzenszwalbPAMI10}
Felzenszwalb, P., Girshick, R., McAllester, D., and Ramanan, D.
\newblock Object detection with discriminatively trained part-based models.
\newblock \emph{PAMI}, 32\penalty0 (9):\penalty0 1627--1645, 2010.

\bibitem[Girshick et~al.(2014)Girshick, Donahue, Darrell, and
  Malik]{GirshickCVPR14}
Girshick, R., Donahue, J., Darrell, T., and Malik, J.
\newblock Rich feature hierarchies for accurate object detection and semantic
  segmentation.
\newblock In \emph{CVPR}, 2014.

\bibitem[Gong et~al.(2012)Gong, Lazebnik, Gordo, and Perronnin]{GongPAMI2012}
Gong, Y., Lazebnik, S., Gordo, A., and Perronnin, F.
\newblock Iterative quantization: A procrustean approach to learning binary
  codes for large-scale image retrieval.
\newblock In \emph{PAMI}, 2012.

\bibitem[Jahrer(2008)]{JahrerCVWW2008}
Jahrer, M., Grabner M. Bischof~H.
\newblock Learned local descriptors for recognition and matching.
\newblock In \emph{Computer Vision Winter Workshop}, 2008.

\bibitem[Jarrett et~al.(2009)Jarrett, Kavukcuoglu, Ranzato, and
  LeCun]{JarrettICCV09}
Jarrett, K., Kavukcuoglu, K., Ranzato, M., and LeCun, Y.
\newblock What is the best multi-stage architecture for object recognition?
\newblock In \emph{ICCV}, 2009.

\bibitem[Kokkinos et~al.(2012)Kokkinos, Bronstein, and Yuille]{KokkinosTR2012}
Kokkinos, I., Bronstein, M., and Yuille, A.
\newblock Dense scale-invariant descriptors for images and surfaces.
\newblock In \emph{INRIA Research Report 7914}, 2012.

\bibitem[Krizhevsky et~al.(2012)Krizhevsky, Sutskever, and
  Hinton]{KrizhevskyNIPS12}
Krizhevsky, A., Sutskever, I., and Hinton, G.
\newblock Imagenet classification with deep convolutional neural networks.
\newblock In \emph{NIPS}, 2012.

\bibitem[Lowe(2004)]{LoweIJCV04}
Lowe, D.
\newblock Distinctive image features from scale-invariant keypoints.
\newblock \emph{IJCV}, 60:\penalty0 91--110, 2004.

\bibitem[Mikolajczyk \& Schmid(2005)Mikolajczyk and Schmid]{MikolajczykPAMI05}
Mikolajczyk, K. and Schmid, C.
\newblock A performance evaluation of local descriptors.
\newblock \emph{PAMI}, 27\penalty0 (10):\penalty0 1615--1630, 2005.

\bibitem[Mobahi et~al.(2009)Mobahi, Collobert, and Weston]{MobahiICML2009}
Mobahi, H., Collobert, R., and Weston, J.
\newblock Deep learning from temporal coherence in video.
\newblock In \emph{ICML}, 2009.

\bibitem[Osendorfer et~al.(2013)Osendorfer, Bayer, Urban, and van~der
  Smagt]{OsendorferICONIP13}
Osendorfer, C., Bayer, J., Urban, S., and van~der Smagt, P.
\newblock Convolutional neural networks learn compact local image descriptors.
\newblock In \emph{ICONIP}, volume 8228. 2013.

\bibitem[Sermanet et~al.(2012)Sermanet, Chintala, and LeCun]{SermanetICPR12}
Sermanet, P., Chintala, S., and LeCun, Y.
\newblock Convolutional neural networks applied to house numbers digit
  classification.
\newblock In \emph{ICPR}, 2012.

\bibitem[Simo-Serra et~al.(2015)Simo-Serra, Torras, and
  Moreno-Noguer]{SimoIJCV2015}
Simo-Serra, E., Torras, C., and Moreno-Noguer, F.
\newblock {DaLI:} deformation and light invariant descriptor.
\newblock \emph{IJCV}, 2015.

\bibitem[Simonyan et~al.(2014)Simonyan, Vedaldi, and Zisserman]{SimonyanPAMI14}
Simonyan, K., Vedaldi, A., and Zisserman, A.
\newblock Learning local feature descriptors using convex optimisation.
\newblock \emph{PAMI}, 2014.

\bibitem[Strecha et~al.(2012)Strecha, Bronstein, Bronstein, and
  Fua]{StrechaPAMI2012}
Strecha, C., Bronstein, A., Bronstein, M., and Fua, P.
\newblock Lda-hash: Improved matching with smaller descriptors.
\newblock In \emph{PAMI}, volume~34, 2012.

\bibitem[Szegedy et~al.(2013)Szegedy, Toshev, and Erhan]{SzegedyNIPS13}
Szegedy, C., Toshev, A., and Erhan, D.
\newblock Deep neural networks for object detection.
\newblock In \emph{NIPS}, 2013.

\bibitem[Trulls et~al.(2013)Trulls, Kokkinos, Sanfeliu, and
  Moreno-Noguer]{TrullsCVPR2013}
Trulls, E., Kokkinos, I., Sanfeliu, A., and Moreno-Noguer, F.
\newblock Dense segmentation-aware descriptors.
\newblock \emph{CVPR}, 2013.

\bibitem[Trzcinski et~al.(2013)Trzcinski, Christoudias, Fua, and
  Lepetit]{TrzcinskiCVPR2013}
Trzcinski, T., Christoudias, M., Fua, P., and Lepetit, V.
\newblock Boosting binary keypoint descriptors.
\newblock In \emph{CVPR}, 2013.

\end{thebibliography}

\newpage

\section{Supplemental Material}
\label{sec:appendix1}

This section contains supplemental material. As we argue in Sec.~\ref{sec:results}, Precision-Recall (PR) curves are the most appropriate metric for our problem; however, we also consider ROC curves and Cumulative Match Curves (CMC).

ROC curves are created by plotting the true positive rate TPR as a function of the true negative rate TNR, where:
\begin{equation}
\mathrm{TPR} =\frac{\mathrm{TP}}{\mathrm{P}} \hspace{1cm} \mathrm{TNR} =1-\frac{\mathrm{FP}}{\mathrm{N}}
\end{equation}

Alternatively, the CMC curve is created by plotting the Rank against the Ratio of correct matches. That is, CMC(k) is the fraction of correct matches that have rank$\leq$k. In particular CMC(1) is the percentage of examples in which the ground truth match is retrieved in the first position.

We report these results for either metric in terms of the curves (plots) and their AUC (tables), for the best-performing iteration.

\subsection{Experiments: (1) Depth and architectures}
\label{sec:appendix1:depth}

\begin{figure}[!p]
	\centering
	\includegraphics[width=0.55\linewidth]{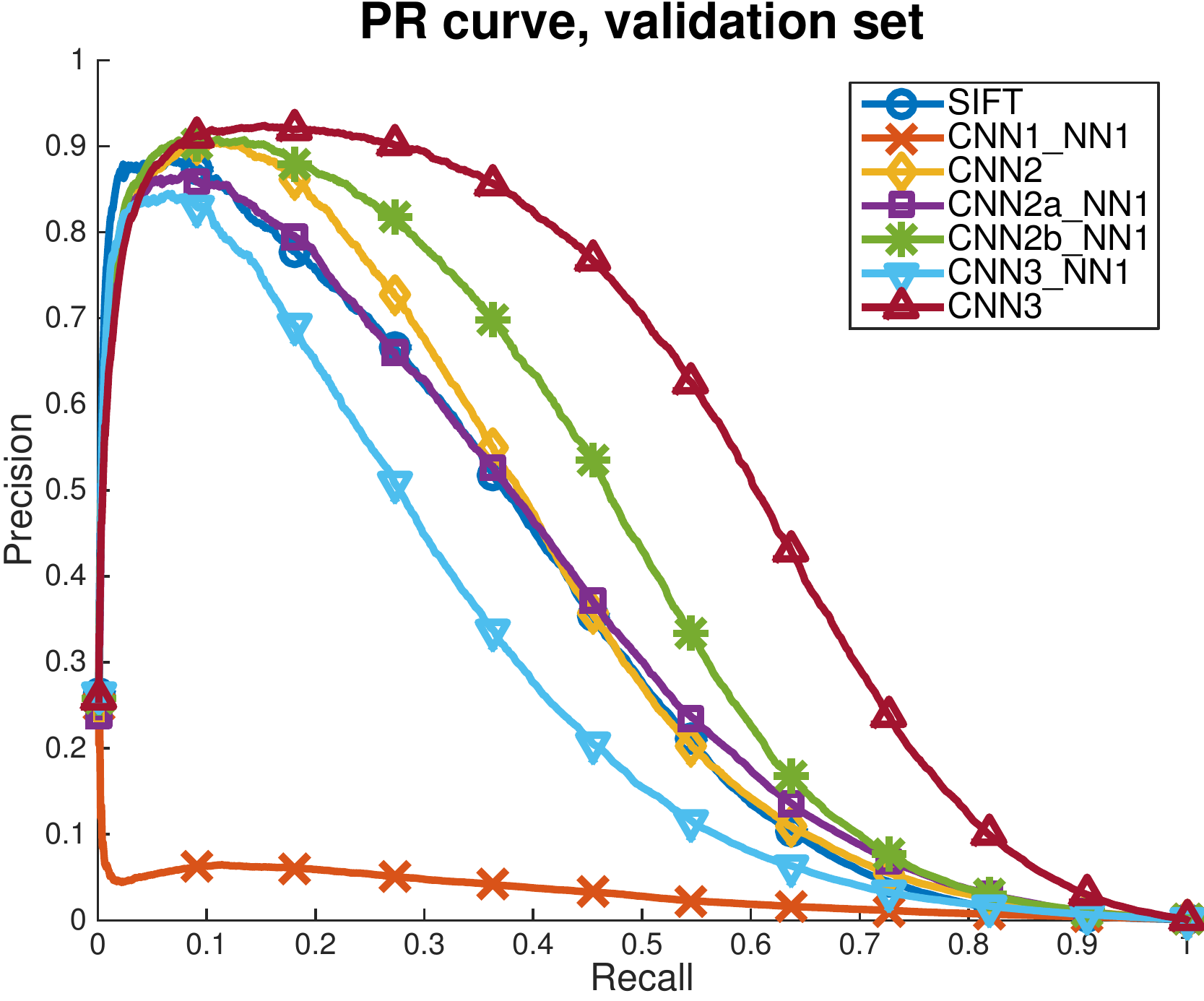}
	\caption{PR curves for the experiments of Sec.~\ref{sec:results:depth} (repeated).}
	\label{fig:appendix:1-pr}
\end{figure}

\begin{figure}[!p]
	\centering
	\includegraphics[width=0.55\linewidth]{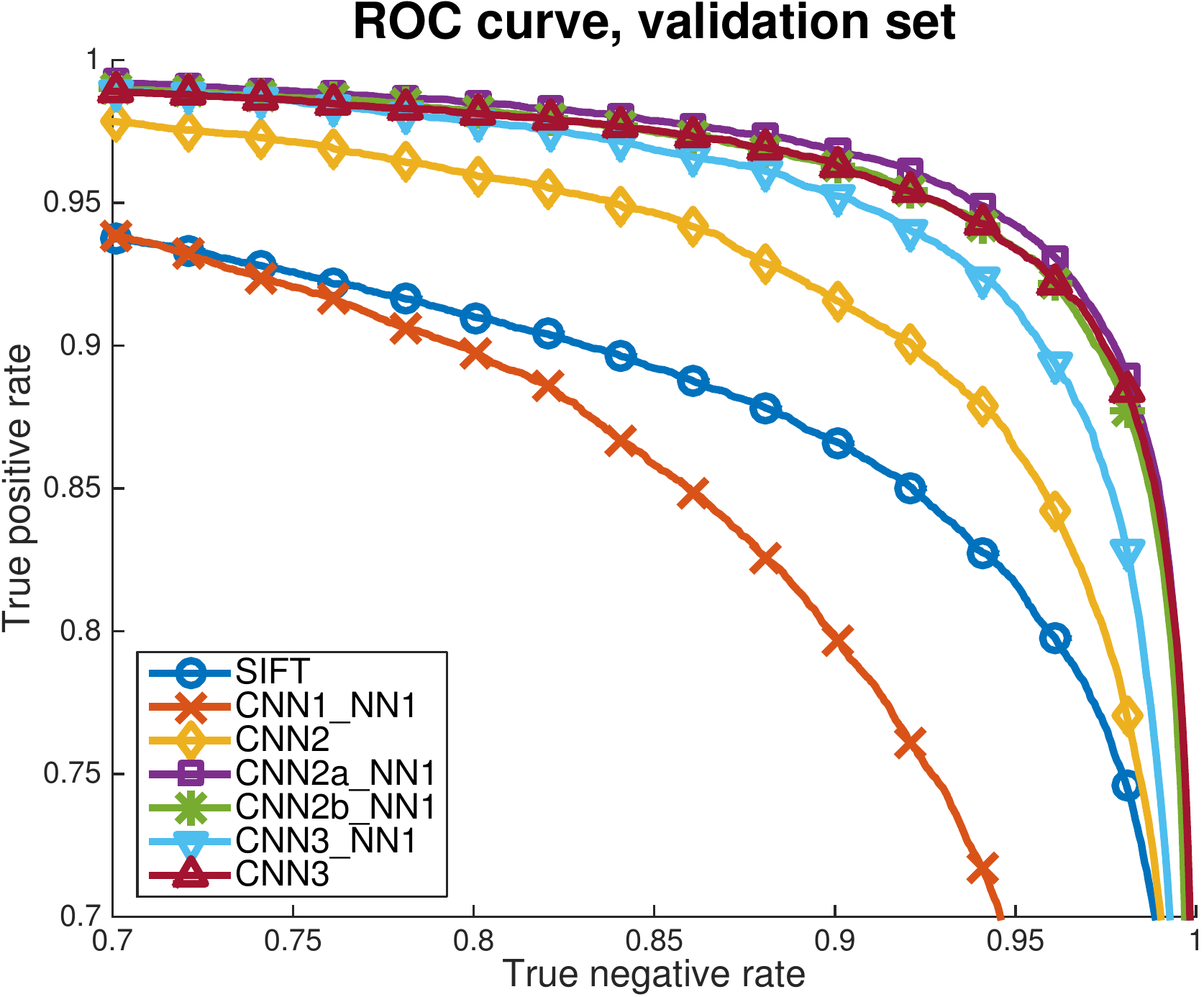}
	\caption{ROC curves for the experiments of Sec.~\ref{sec:results:depth}.}
	\label{fig:appendix:1-roc}
\end{figure}

\begin{figure}[!p]
	\centering
	\includegraphics[width=0.55\linewidth]{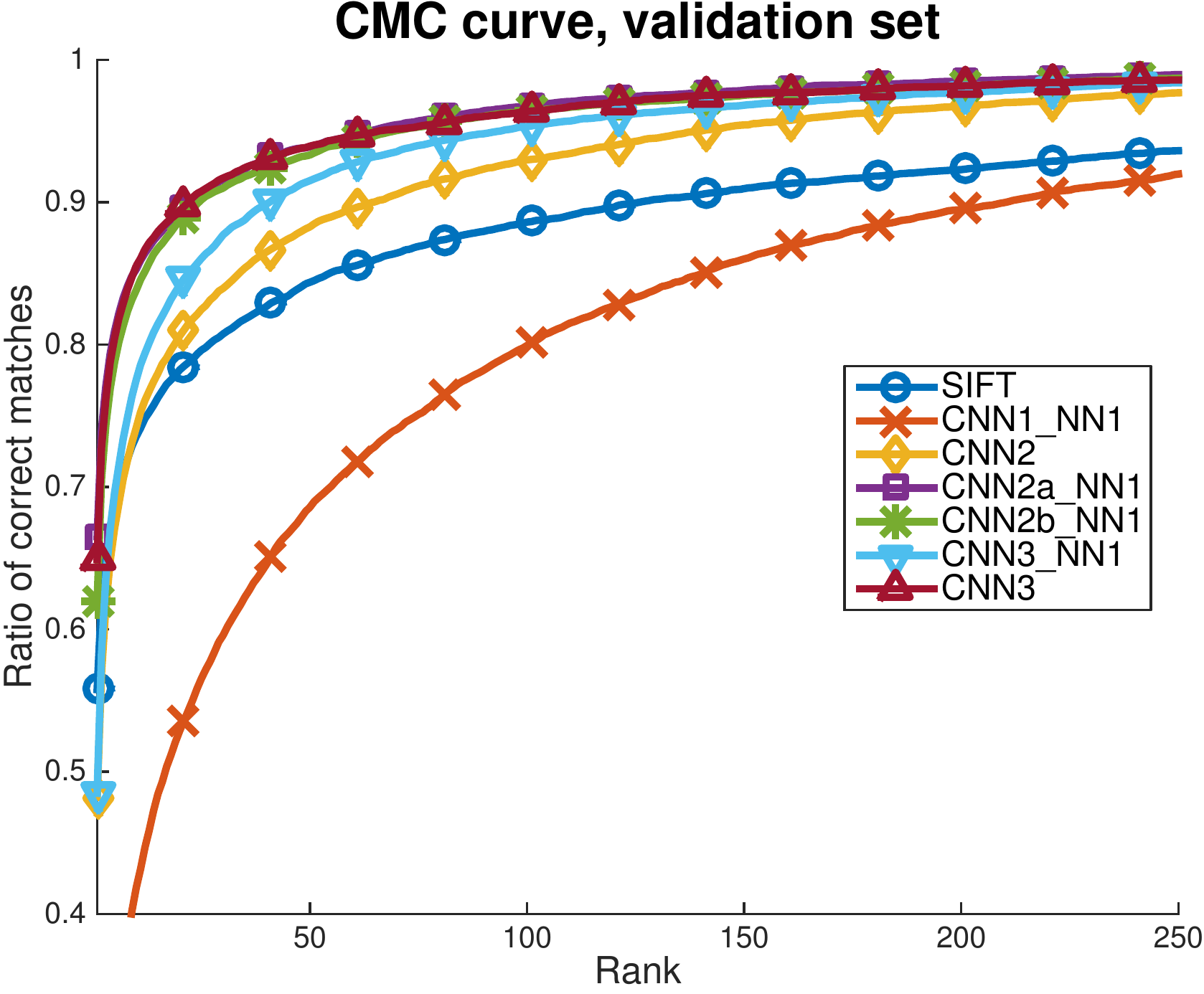}
	\caption{CMC curves for the experiments of Sec.~\ref{sec:results:depth}.}
	\label{fig:appendix:1-cmc}
\end{figure}

We extend the results of Sec.~\ref{sec:results:depth}, which are summarized in Table~\ref{tbl:appendix:1}. Figs.~\ref{fig:appendix:1-pr}, \ref{fig:appendix:1-roc} and \ref{fig:appendix:1-cmc} show the PR, ROC and CMC curves respectively.

\begin{table}[!h]
	\centering
	\renewcommand{\arraystretch}{1.2}
	\begin{tabular}{lcccc}
	\toprule
	Architecture & Parameters & PR AUC & ROC AUC & CMC AUC \\
	\midrule
	SIFT & --- & 0.361 & 0.944 & 0.953 \\
	CNN1\_NN1 & 68,352 & 0.032 & 0.929 & 0.929 \\
	CNN2 & 27,776 & 0.379 & 0.971 & 0.975 \\
	CNN2a\_NN1 & 145,088 & 0.370 & {\bf 0.987} & {\bf 0.988} \\
	CNN2b\_NN1 & 48,576 & 0.439 & 0.985 & 0.986 \\
	CNN3\_NN1 & 62,784 & 0.289 & 0.980 & 0.982 \\
	CNN3 & 46,272 & {\bf 0.558} & 0.986 & 0.987 \\
	\bottomrule
	\end{tabular}
	\caption{Extended table for the experiments of Sec.~\ref{sec:results:units}.}
	\label{tbl:appendix:1}
\end{table}

\clearpage
\subsection{Experiments: (2) Hidden units, normalization and pooling}
\label{sec:appendix1:units}

\begin{figure}[!p]
	\centering
	\includegraphics[width=0.55\linewidth]{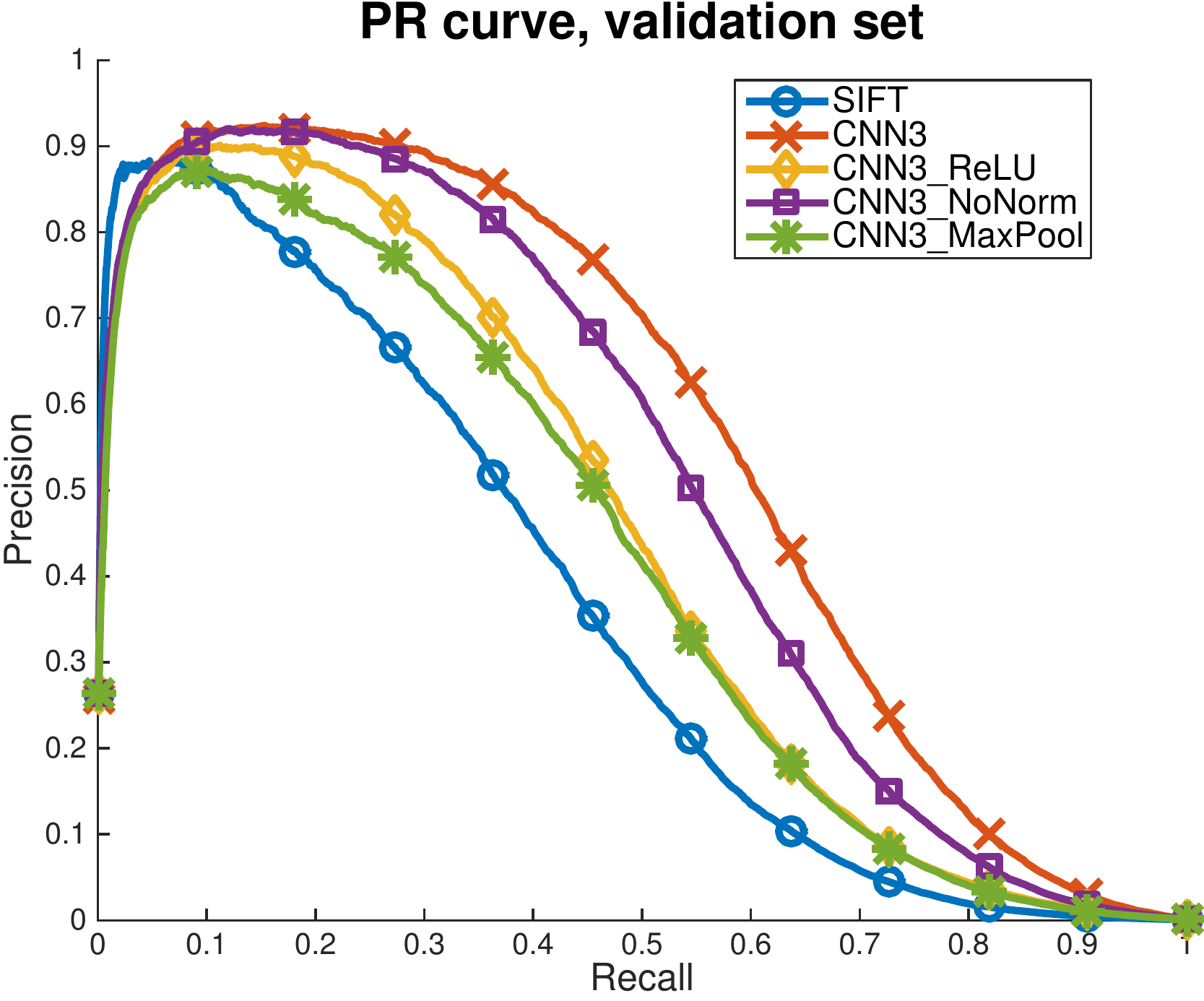}
	\caption{PR curves for the experiments of Sec.~\ref{sec:results:units} (repeated).}
	\label{fig:appendix:2-pr}
\end{figure}

\begin{figure}[!p]
	\centering
	\includegraphics[width=0.55\linewidth]{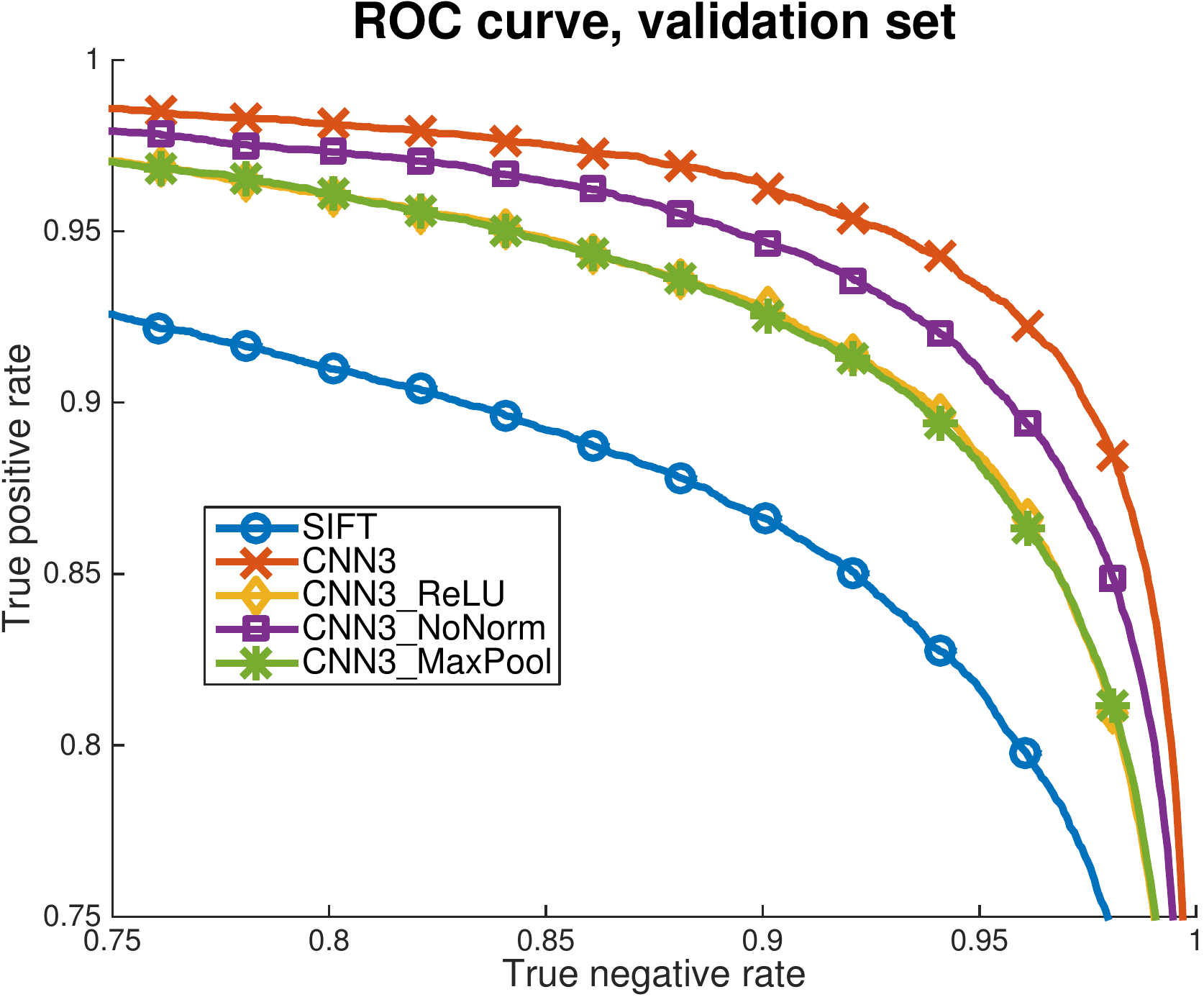}
	\caption{ROC curves for the experiments of Sec.~\ref{sec:results:units}.}
	\label{fig:appendix:2-roc}
\end{figure}

\begin{figure}[!p]
	\centering
	\includegraphics[width=0.55\linewidth]{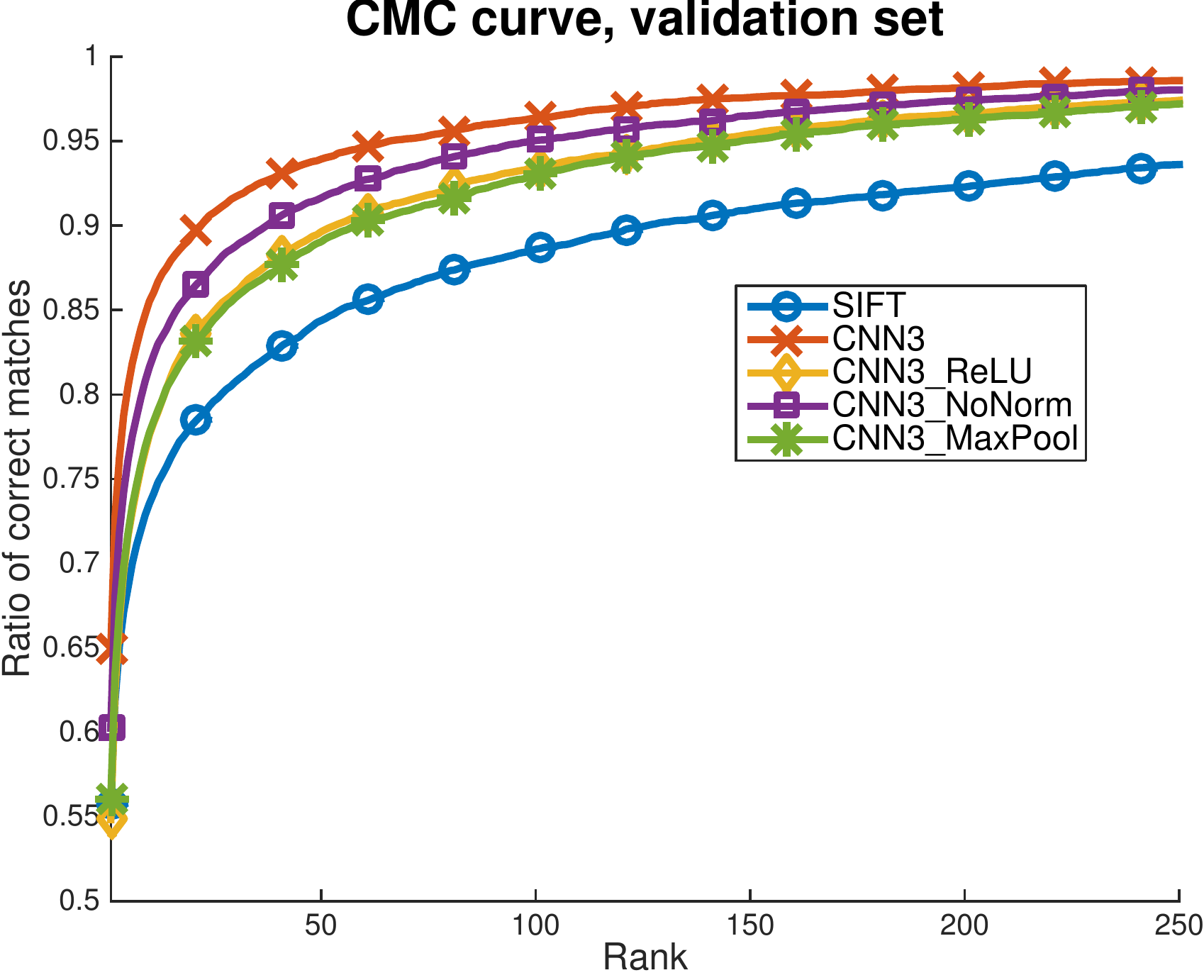}
	\caption{CMC curves for the experiments of Sec.~\ref{sec:results:units}.}
	\label{fig:appendix:2-cmc}
\end{figure}

We extend the results of Sec.~\ref{sec:results:units}, which are summarized in Table~\ref{tbl:appendix:2}. Figs.~\ref{fig:appendix:2-pr}, \ref{fig:appendix:2-roc} and \ref{fig:appendix:2-cmc} show the PR, ROC and CMC curves respectively.

\begin{table}[!h]
	\centering
	\renewcommand{\arraystretch}{1.2}
	\begin{tabular}{lccc}
	\toprule
	Architecture & PR AUC & ROC AUC & CMC AUC \\
	\midrule
	SIFT & 0.361 & 0.944 & 0.953 \\
	CNN3 & {\bf 0.558} & {\bf 0.986} & {\bf 0.987} \\
	CNN3 ReLU & 0.442 & 0.973 & 0.976 \\
	CNN3 No Norm & 0.511 & 0.980 & 0.982 \\
	CNN3 MaxPool & 0.420 & 0.973 & 0.975 \\
	\bottomrule
	\end{tabular}
	\caption{Extended table for the experiments of Sec.~\ref{sec:results:units}.}
	\label{tbl:appendix:2}
\end{table}

\clearpage
\subsection{Experiments: (3) Fracking}
\label{sec:appendix1:fracking}

\begin{figure}[!p]
	\centering
	\includegraphics[width=0.55\linewidth]{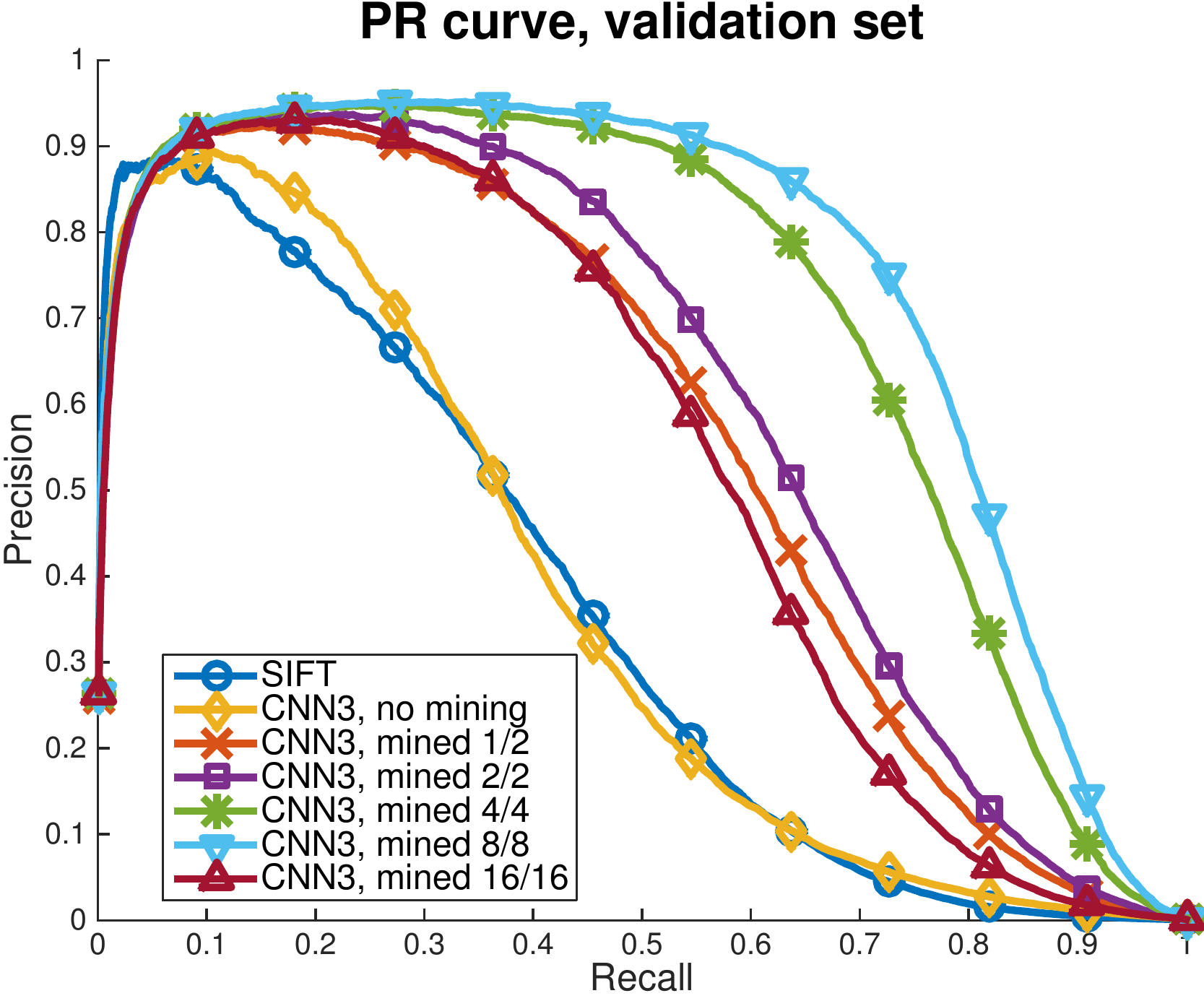}
	\caption{PR curves for the experiments of Sec.~\ref{sec:results:fracking} (repeated).}
	\label{fig:appendix:3-pr}
\end{figure}

\begin{figure}[!p]
	\centering
	\includegraphics[width=0.55\linewidth]{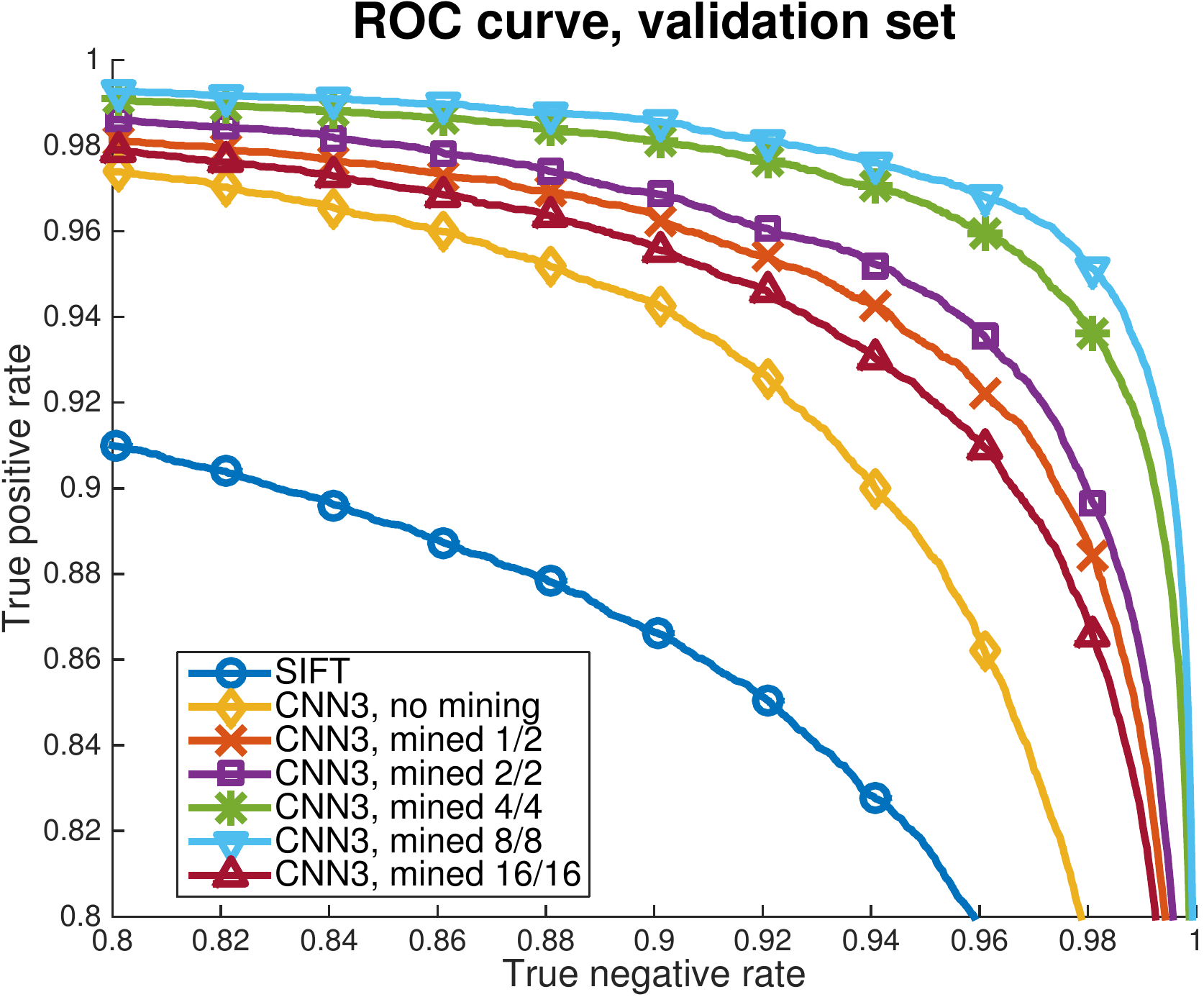}
	\caption{ROC curves for the experiments of Sec.~\ref{sec:results:fracking}.}
	\label{fig:appendix:3-roc}
\end{figure}

\begin{figure}[!p]
	\centering
	\includegraphics[width=0.55\linewidth]{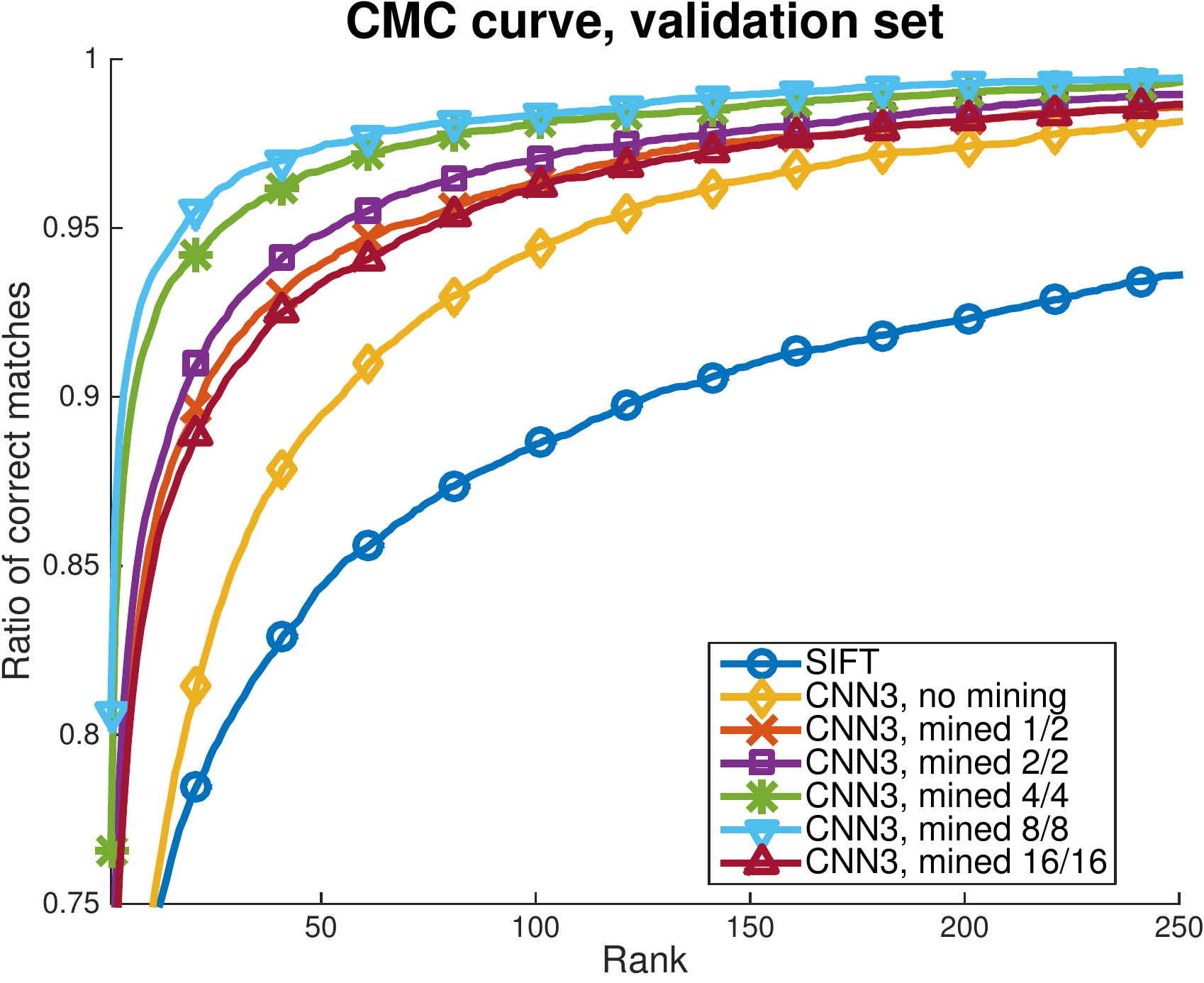}
	\caption{CMC curves for the experiments of Sec.~\ref{sec:results:fracking}.}
	\label{fig:appendix:3-cmc}
\end{figure}

We extend the results of Sec.~\ref{sec:results:fracking}, which are summarized in Table~\ref{tbl:appendix:3}. Figs.~\ref{fig:appendix:3-pr}, \ref{fig:appendix:3-roc} and \ref{fig:appendix:3-cmc} show the PR, ROC and CMC curves respectively.

\begin{table}[!h]
\centering
	\begin{tabular}{ccccc}
	\toprule
	$\frac{B_P}{B_P^M}$ & $\frac{B_N}{B_N^M}$ & PR AUC & ROC AUC & CMC AUC \\
	\midrule
	1 & 1 & 0.366 & 0.977 & 0.979 \\
	1 & 2 & 0.558 & 0.986 & 0.987 \\
	2 & 2 & 0.596 & 0.988 & 0.989 \\
	4 & 4 & 0.703 & 0.993 & 0.993 \\
	8 & 8 & {\bf 0.746} & {\bf 0.994} & {\bf 0.994} \\
	16 & 16 & 0.538 & 0.983 & 0.986 \\
	\bottomrule
	\end{tabular}
	\caption{Extended table for the experiments of Sec.~\ref{sec:results:fracking}.}
	\label{tbl:appendix:3}
\end{table}

\clearpage
\subsection{Experiments: (4) Number of filters and descriptor dimension}
\label{sec:appendix1:filter-dims}

\begin{figure}[!p]
	\centering
	\includegraphics[width=0.55\linewidth]{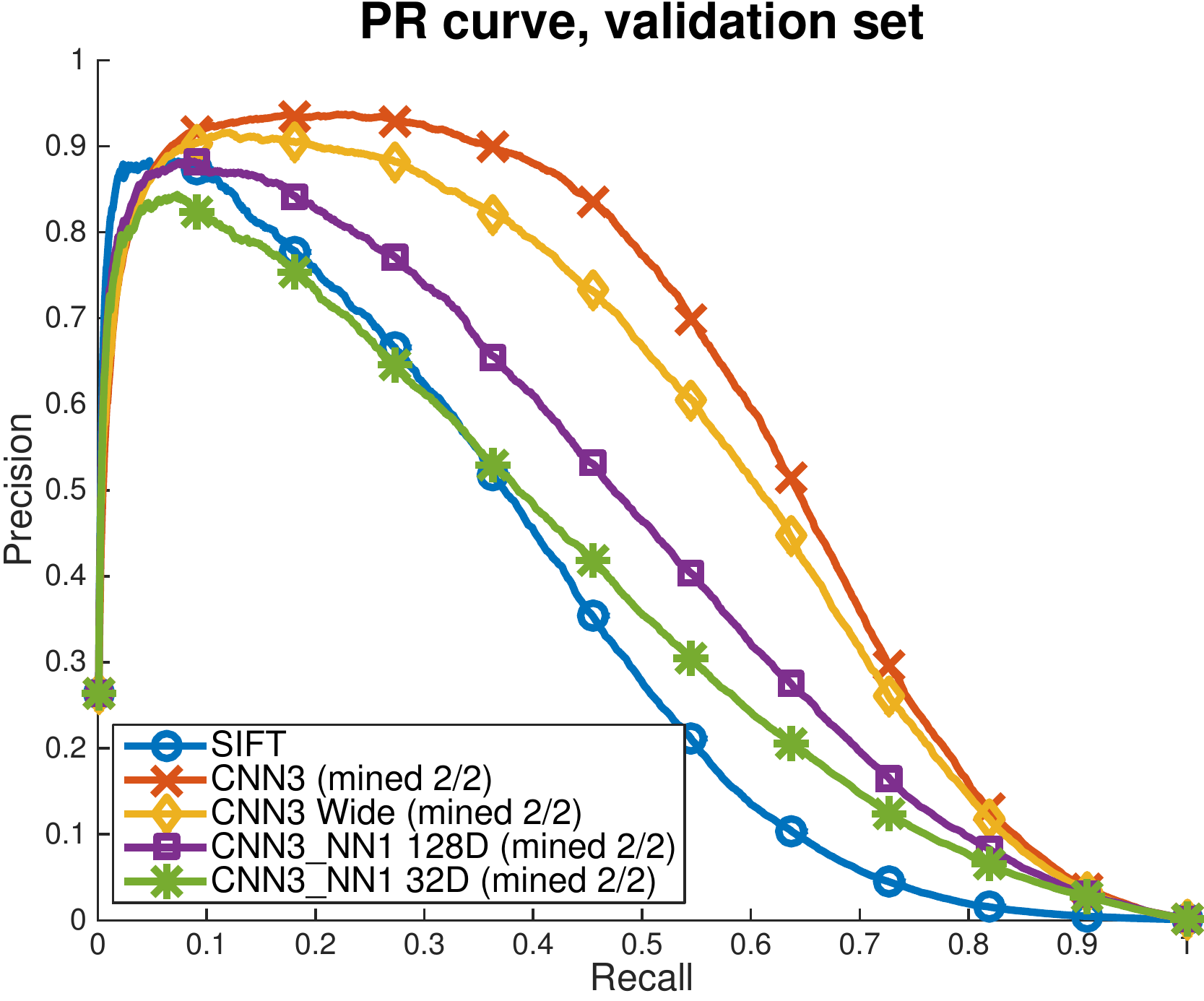}
	\caption{PR curves for the experiments of Sec.~\ref{sec:results:filter-dims} (repeated).}
	\label{fig:appendix:4-pr}
\end{figure}

\begin{figure}[!p]
	\centering
	\includegraphics[width=0.55\linewidth]{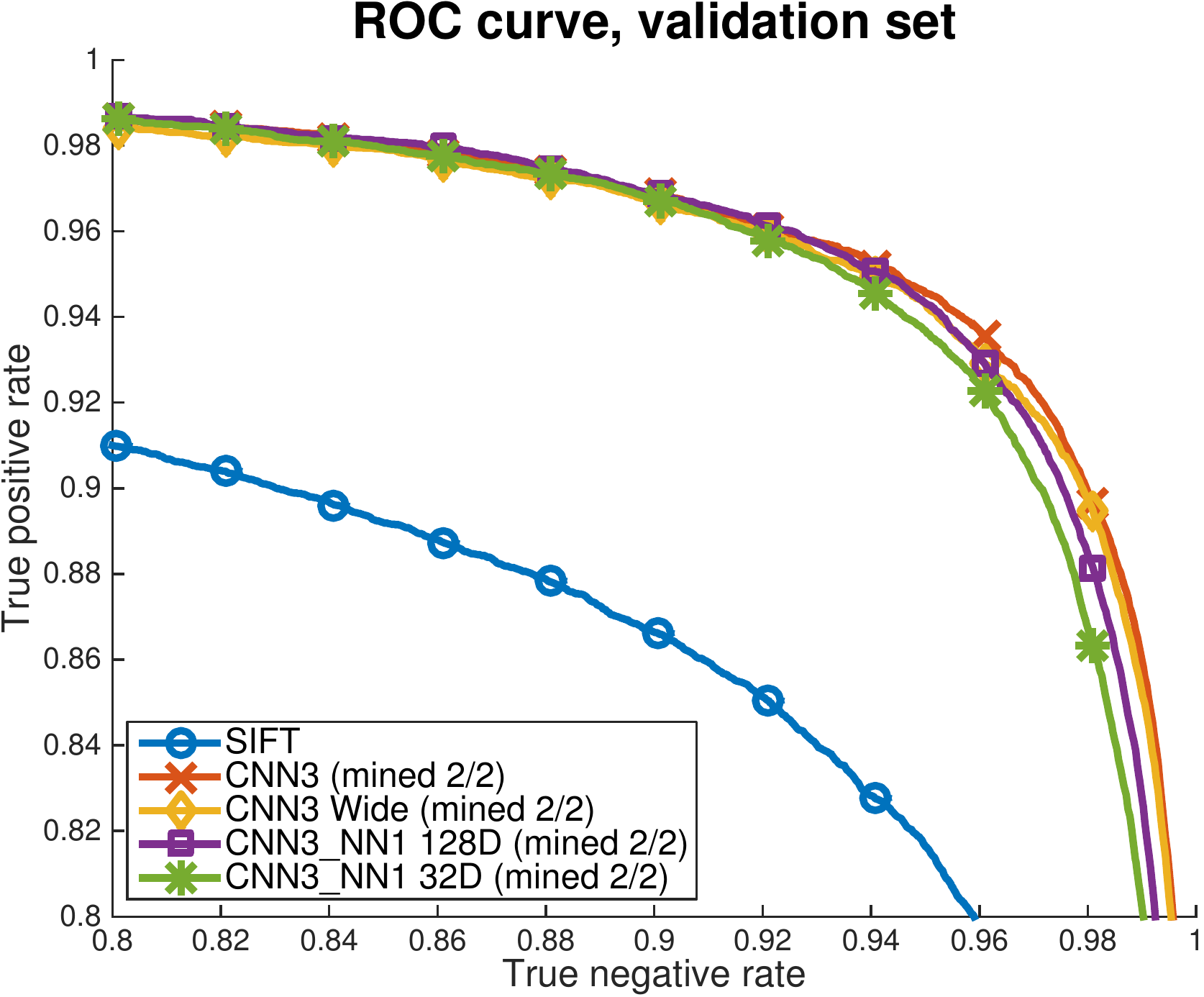}
	\caption{ROC curves for the experiments of Sec.~\ref{sec:results:filter-dims}.}
	\label{fig:appendix:4-roc}
\end{figure}

\begin{figure}[!p]
	\centering
	\includegraphics[width=0.55\linewidth]{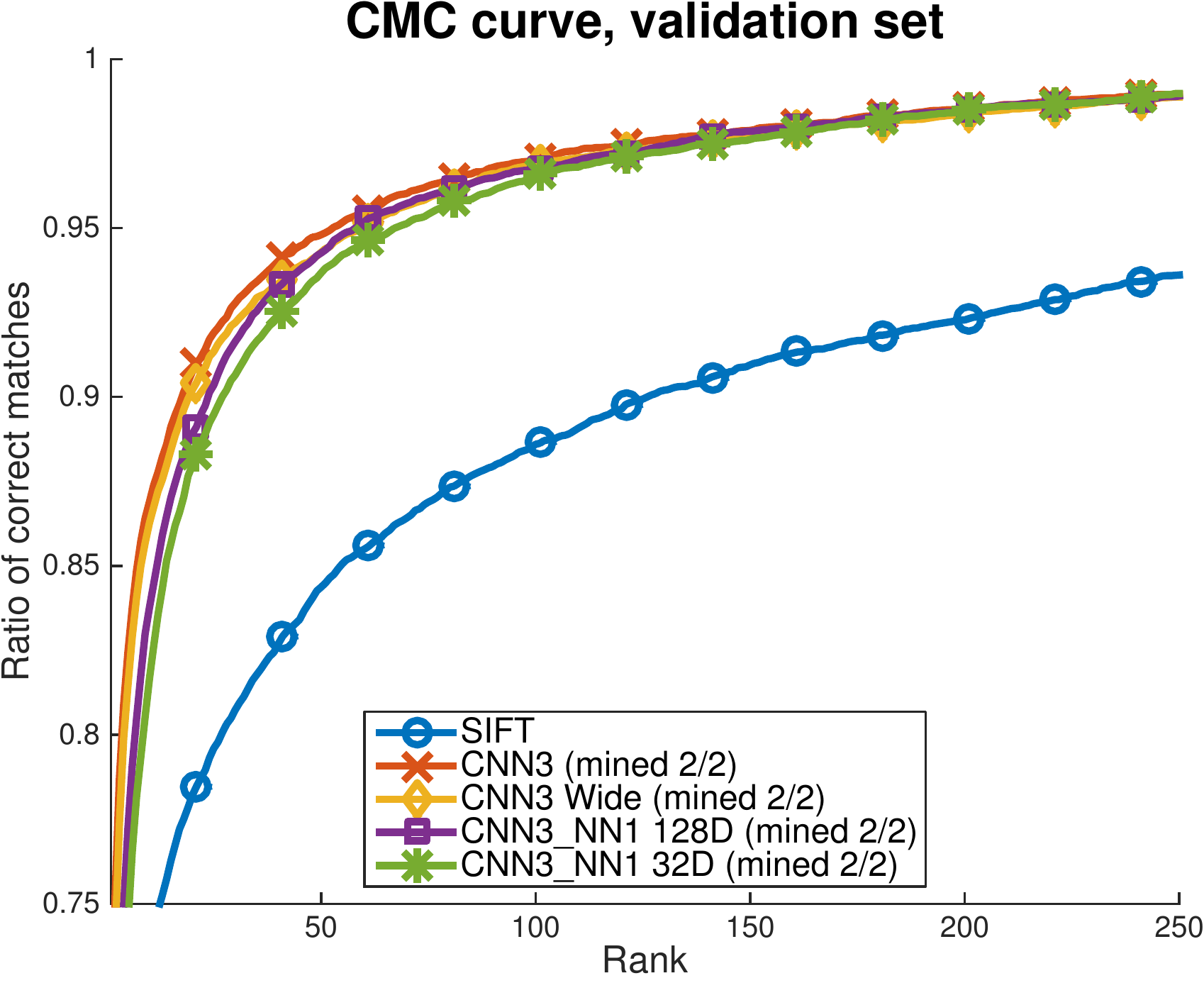}
	\caption{CMC curves for the experiments of Sec.~\ref{sec:results:filter-dims}.}
	\label{fig:appendix:4-cmc}
\end{figure}

We extend the results of Sec.~\ref{sec:results:filter-dims}, which are summarized in Table~\ref{tbl:appendix:4}. Figs.~\ref{fig:appendix:4-pr}, \ref{fig:appendix:4-roc} and \ref{fig:appendix:4-cmc} show the PR, ROC and CMC curves respectively.

\begin{table}[!h]
\centering
	\renewcommand{\arraystretch}{1.2}
	\begin{tabular}{lccccc}
	\toprule
	Architecture & Output & Parameters & PR AUC & ROC AUC & CMC AUC \\
	\midrule
	SIFT & 128D & --- & 0.361 & 0.944 & 0.953 \\
	CNN3 & 128D & 46,272 & {\bf 0.596} & {\bf 0.988} & {\bf 0.989} \\
	CNN3 Wide & 128D & 110,496 & 0.552 & 0.987 & 0.988 \\
	CNN3\_NN1 & 128D & 62,784 & 0.456 & 0.988 & 0.988 \\
	CNN3\_NN1 & 32D & 50,400 & 0.389 & 0.986 & 0.987 \\
	\bottomrule
	\end{tabular}
	\caption{Extended table for the experiments of Sec.~\ref{sec:results:filter-dims}.}
	\label{tbl:appendix:4}
\end{table}

\clearpage
\subsection{Experiments: (5) Generalization}
\label{sec:appendix1:generalization}

\begin{figure}[!p]
	\centering
	\includegraphics[width=0.55\linewidth]{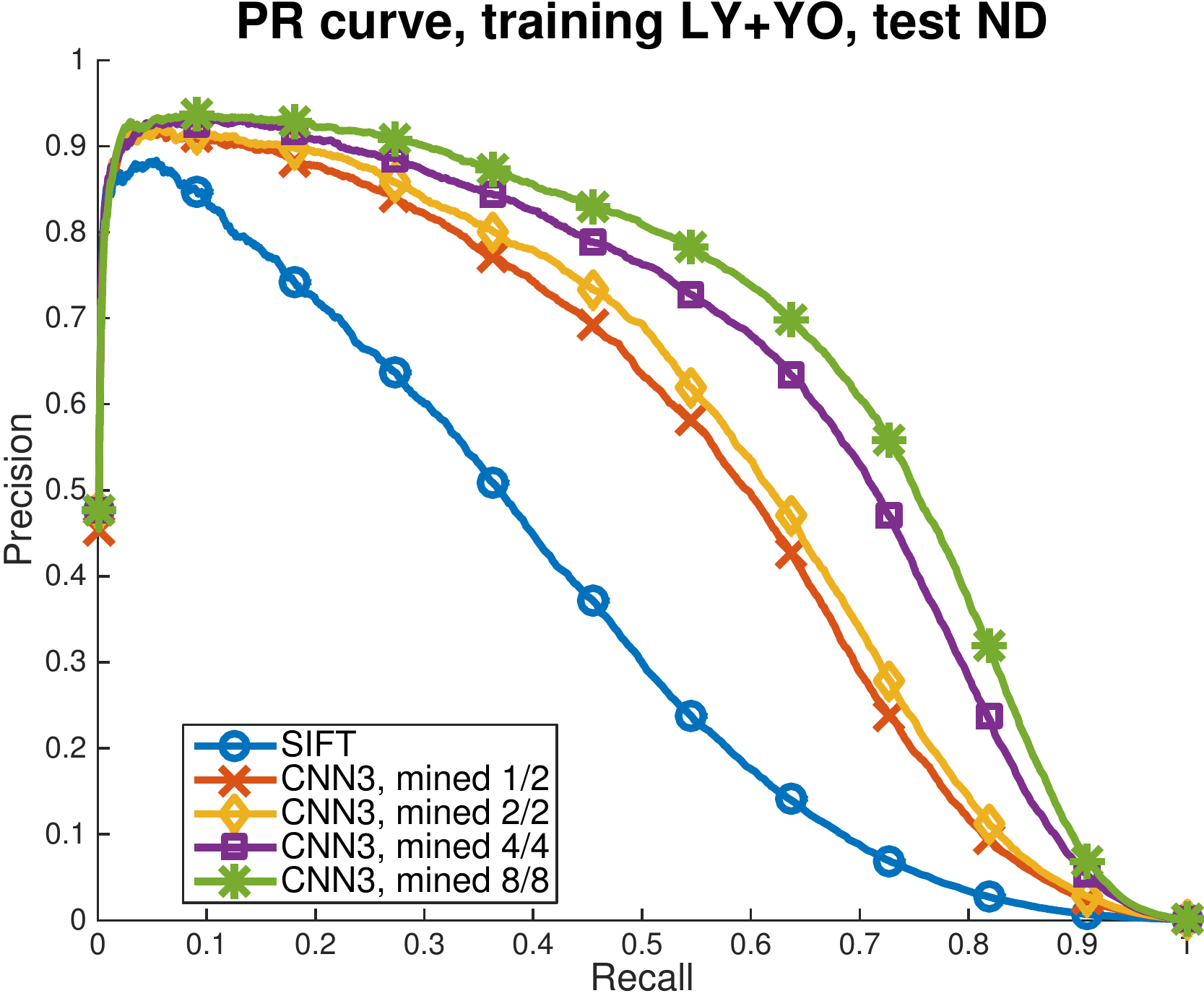}
	\caption{Generalized results in PR, first split.}
	\label{fig:appendix:5-pr1}
\end{figure}

\begin{figure}[!p]
	\centering
	\includegraphics[width=0.55\linewidth]{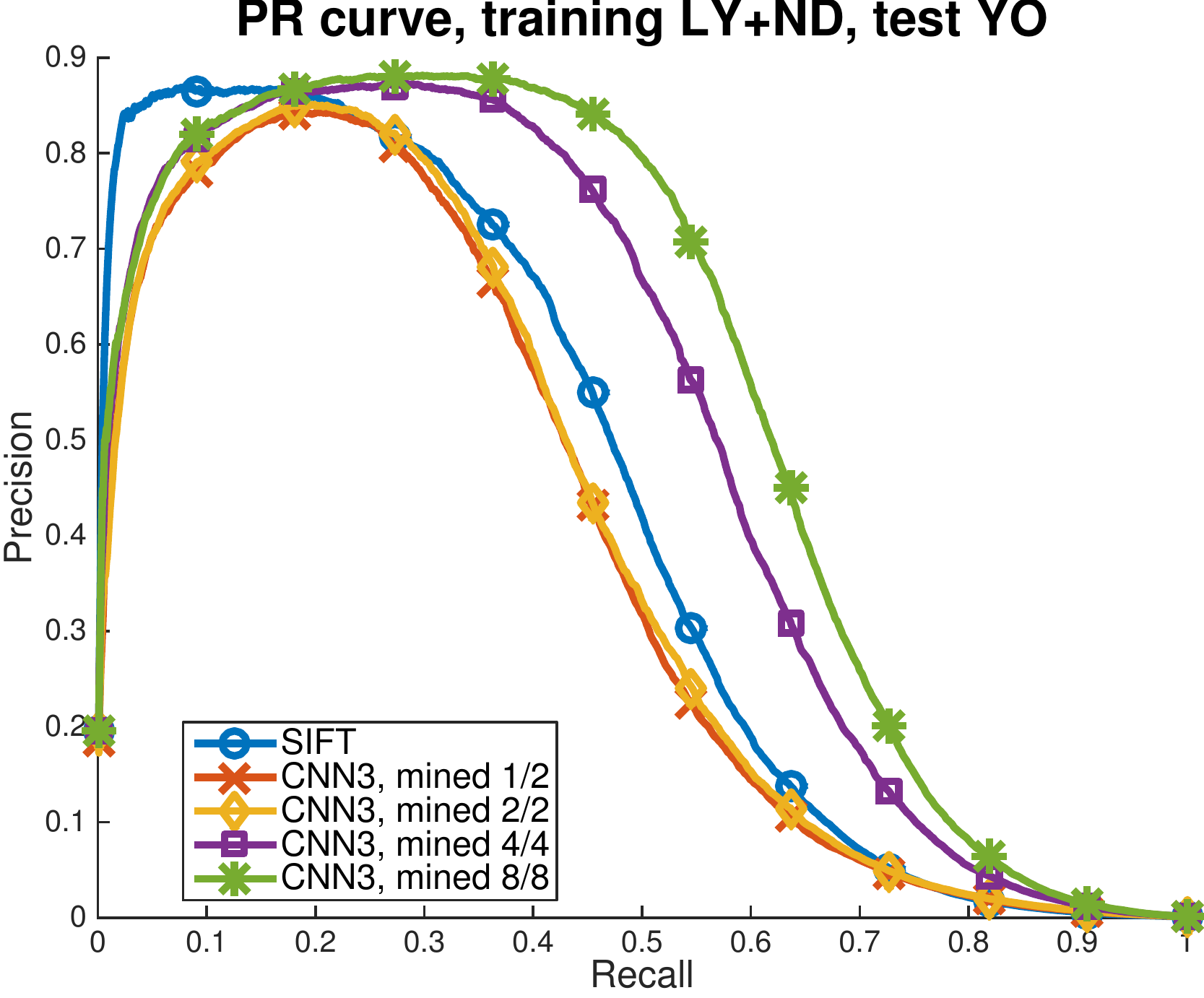}
	\caption{Generalized results in PR, second split.}
	\label{fig:appendix:5-pr2}
\end{figure}

\begin{figure}[!p]
	\centering
	\includegraphics[width=0.55\linewidth]{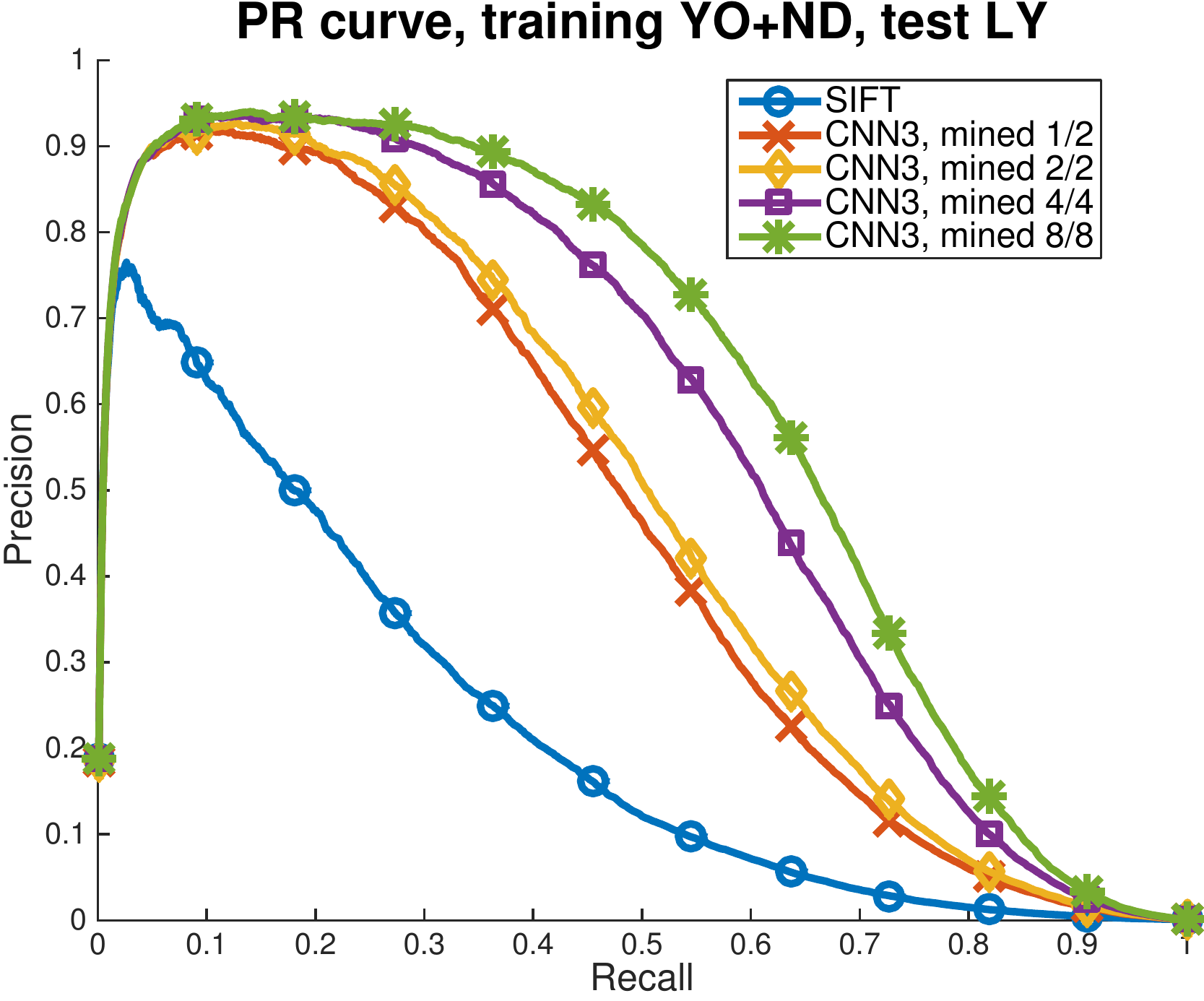}
	\caption{Generalized results in PR, third split.}
	\label{fig:appendix:5-pr3}
\end{figure}

\begin{figure}[!p]
	\centering
	\includegraphics[width=0.55\linewidth]{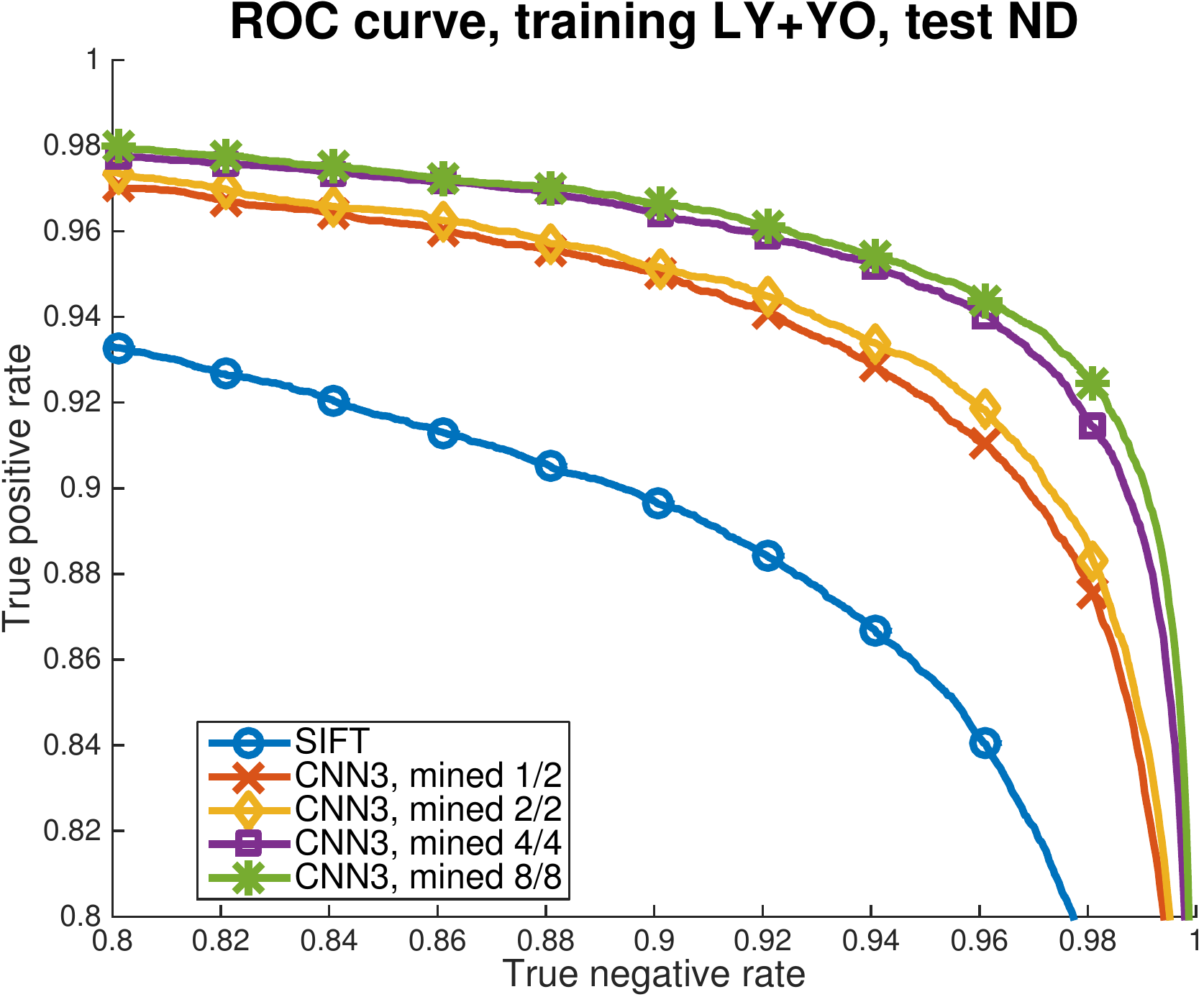}
	\caption{Generalized results in ROC, first split.}
	\label{fig:appendix:5-roc1}
\end{figure}

\begin{figure}[!p]
	\centering
	\includegraphics[width=0.55\linewidth]{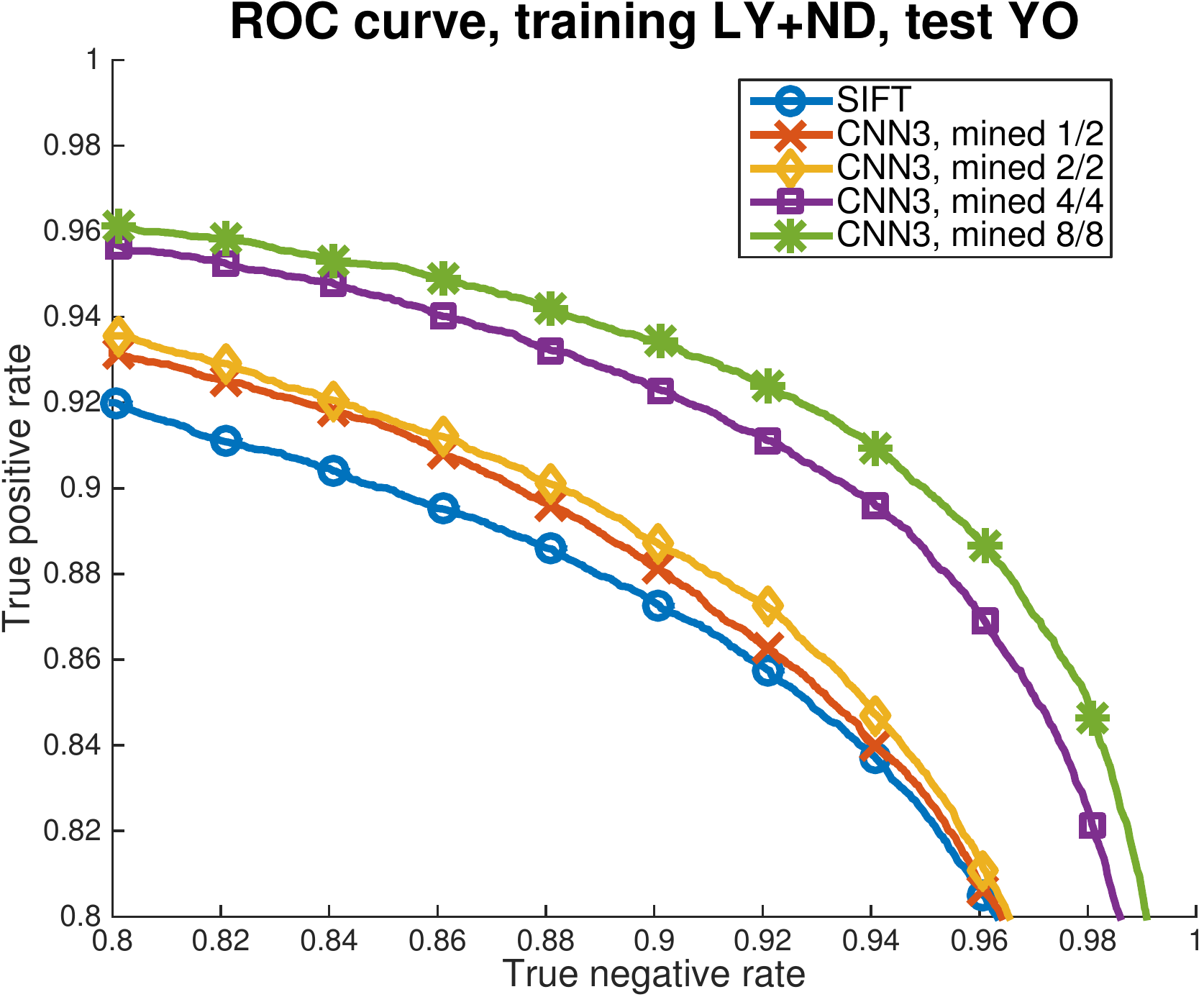}
	\caption{Generalized results in ROC, second split.}
	\label{fig:appendix:5-roc2}
\end{figure}

\begin{figure}[!p]
	\centering
	\includegraphics[width=0.55\linewidth]{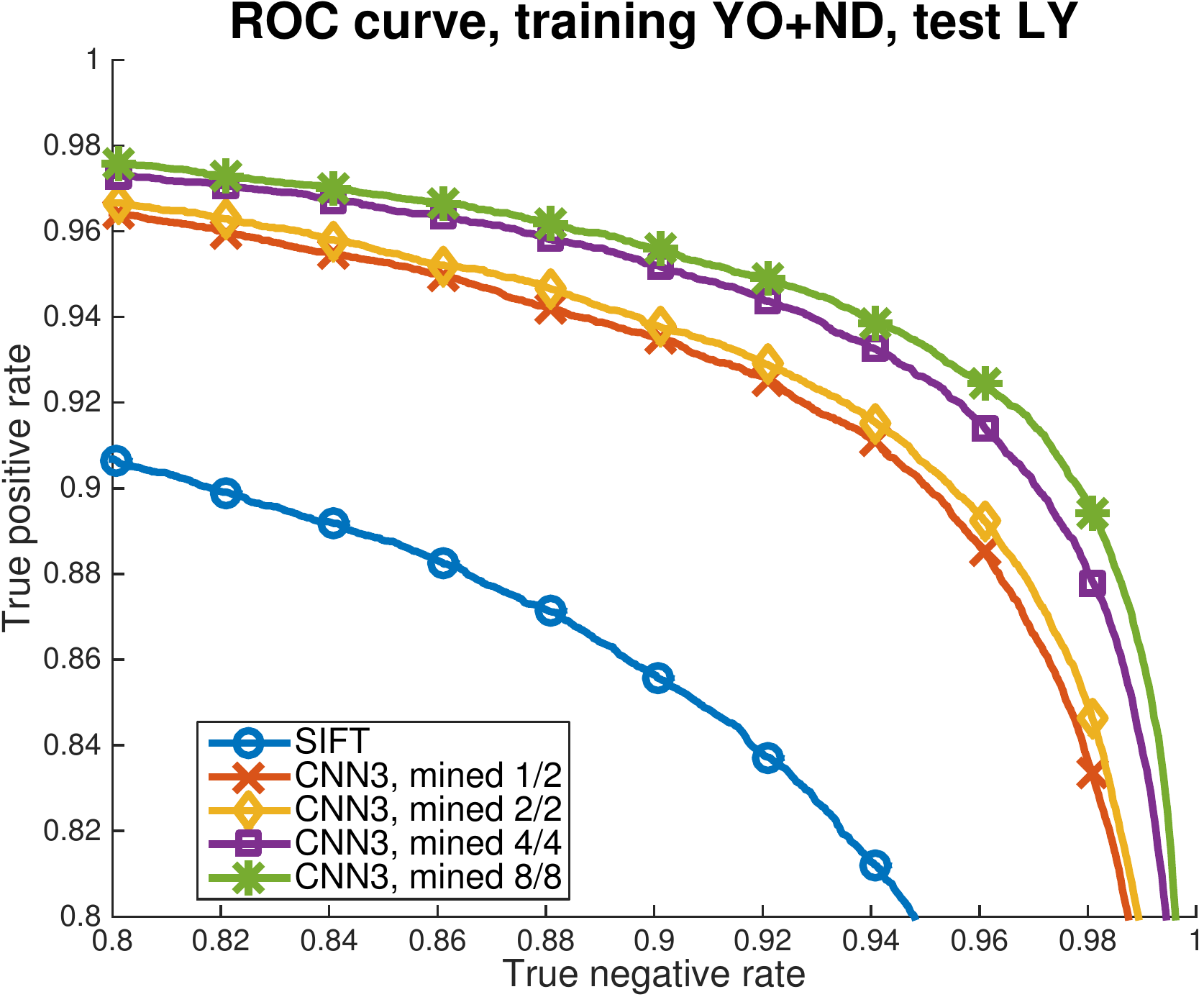}
	\caption{Generalized results in ROC, third split.}
	\label{fig:appendix:5-roc3}
\end{figure}

\begin{figure}[!p]
	\centering
	\includegraphics[width=0.55\linewidth]{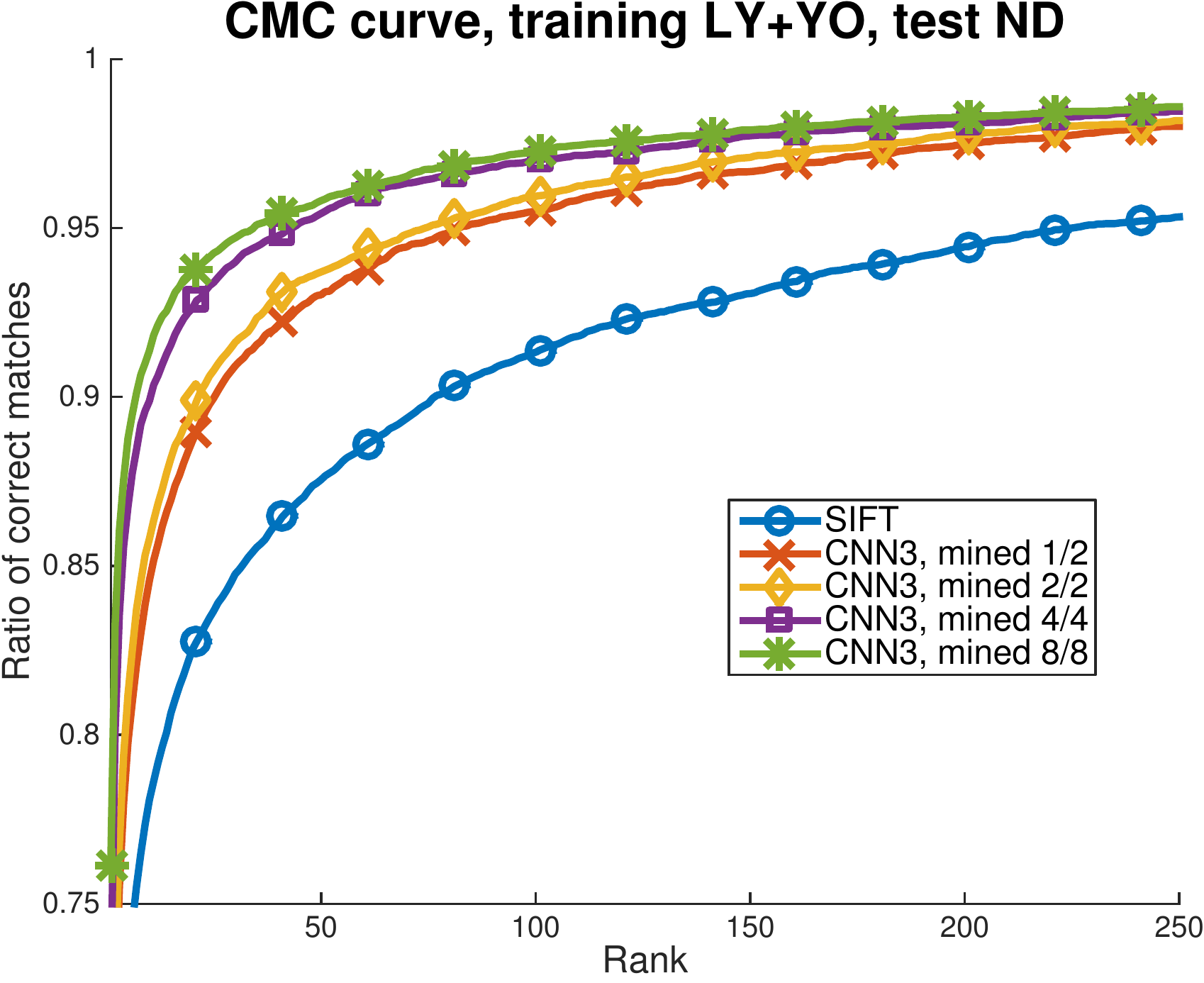}
	\caption{Generalized results in CMC, first split.}
	\label{fig:appendix:5-cmc1}
\end{figure}

\begin{figure}[!p]
	\centering
	\includegraphics[width=0.55\linewidth]{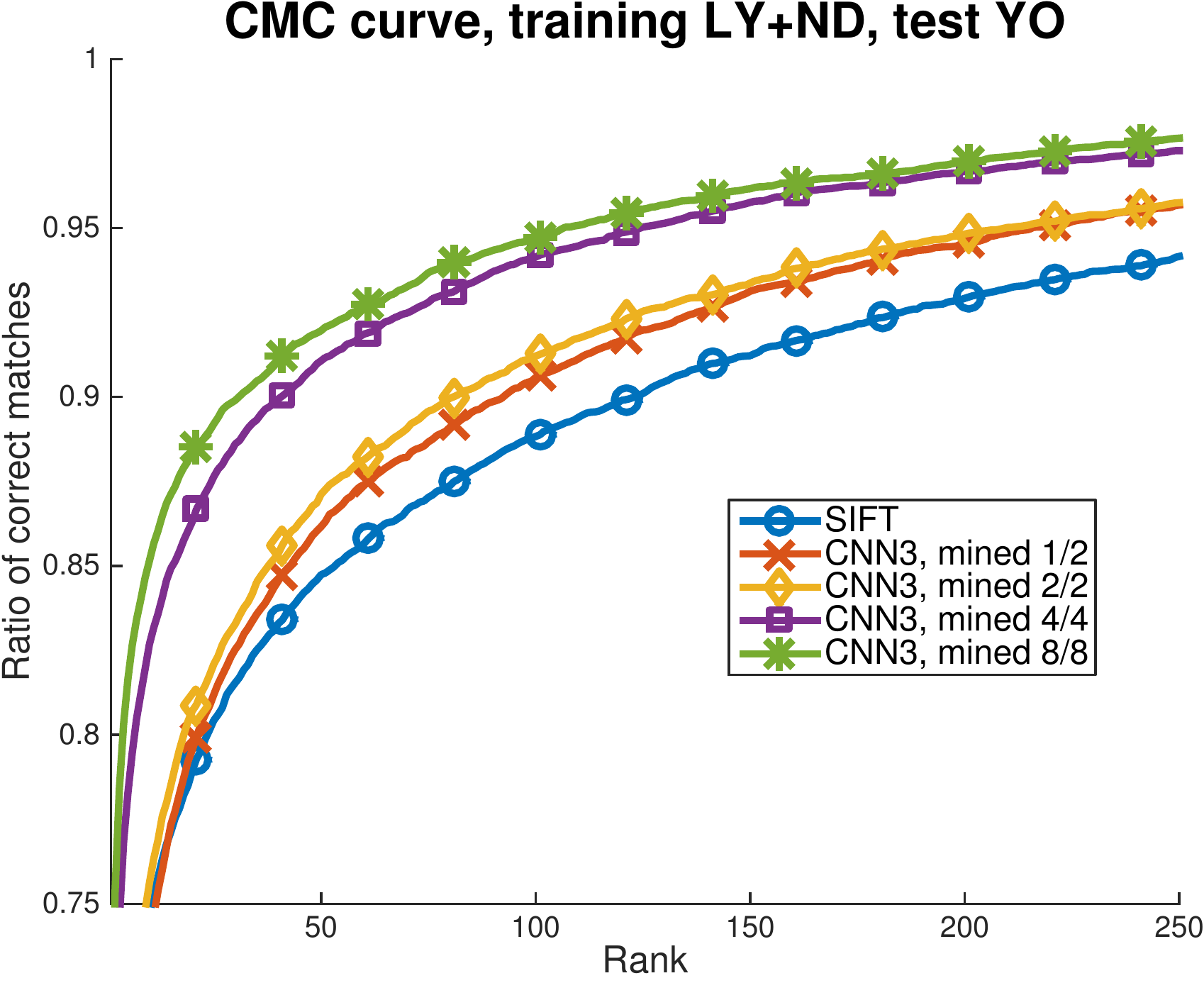}
	\caption{Generalized results in CMC, second split.}
	\label{fig:appendix:5-cmc2}
\end{figure}

\begin{figure}[!p]
	\centering
	\includegraphics[width=0.55\linewidth]{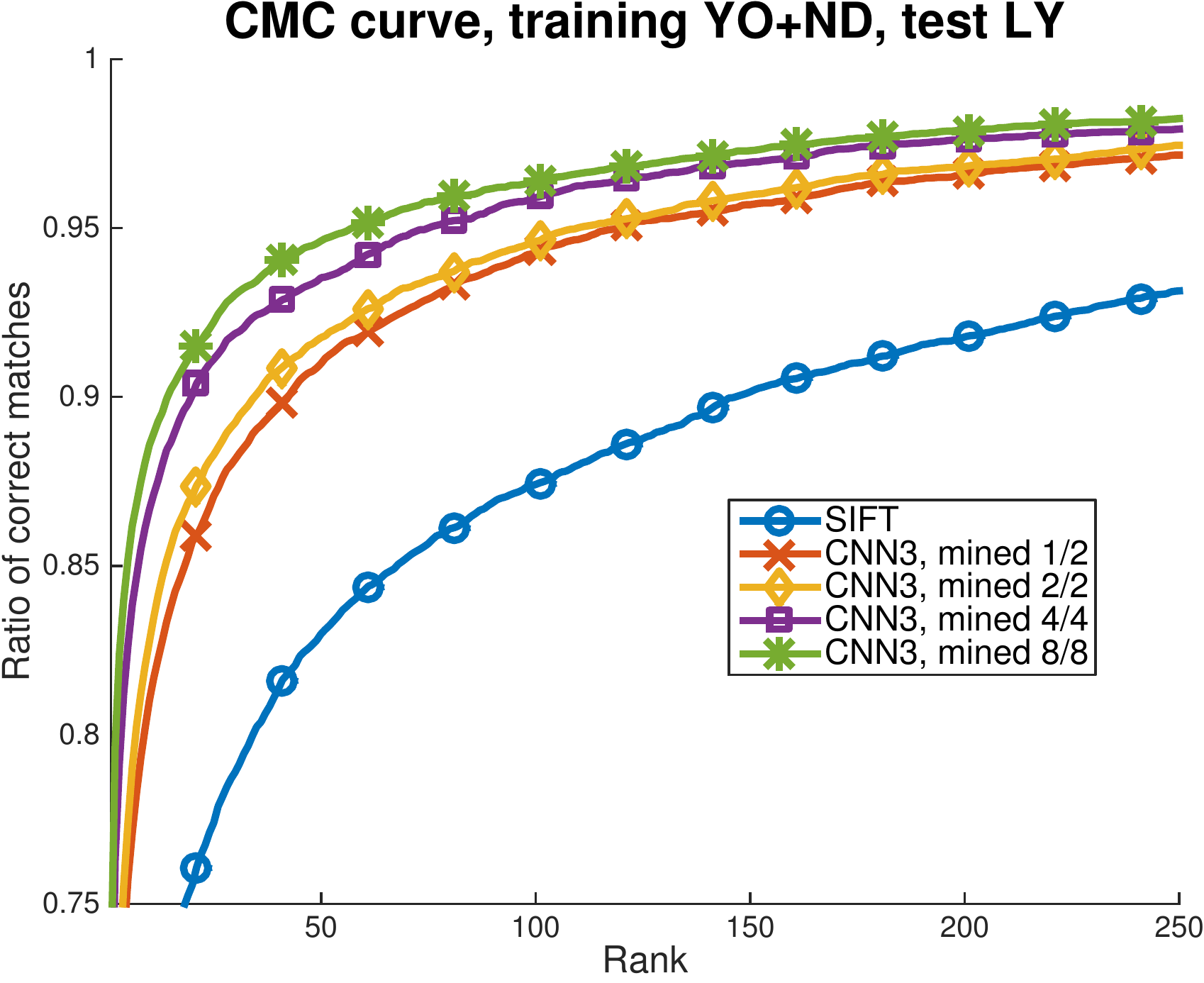}
	\caption{Generalized results in CMC, third split.}
	\label{fig:appendix:5-cmc3}
\end{figure}

In this section we extend the results of Sec.~\ref{sec:results:generalization}. We summarize the results over three different dataset splits, each with ten test folds of 10,000 randomly sampled positives and 1,000 randomly sampled negatives. We show the PR results in Tables~\ref{tbl:appendix:5-pr1}-\ref{tbl:appendix:5-pr3}, and Figs.~\ref{fig:appendix:5-pr1}-\ref{fig:appendix:5-pr3}, the ROC results in Tables~\ref{tbl:appendix:5-roc1}-\ref{tbl:appendix:5-roc3}, and Figs.~\ref{fig:appendix:5-roc1}-\ref{fig:appendix:5-roc3}, and the CMC results in Tables~\ref{tbl:appendix:5-cmc1}-\ref{tbl:appendix:5-cmc3}, and Figs.~\ref{fig:appendix:5-cmc1}-\ref{fig:appendix:5-cmc3}.


\begin{table}[!h]
\centering
\def\arraystretch{1.1}
\setlength{\tabcolsep}{4pt}
\begin{tabular}{ccccccccccccc}
	\toprule
	& \multicolumn{10}{c}{{\bf Precision-Recall AUC, Train: LY+YOS, Test: ND (10 folds)}} & \\
	\hline
	Model & F1 & F2 & F3 & F4 & F5 & F6 & F7 & F8 & F9 & F10 & Avg. & Increase \\
	\hline \hline
	SIFT           & .364 & .352 & .345 & .343 & .349 & .350 & .350 & .351 & .341 & .348 & .349 & --- \\
	CNN3, mine-1/2 & .535 & .527 & .538 & .537 & .548 & .529 & .537 & .535 & .530 & .529 & .535 & 53.3\% \\
	CNN3, mine-2/2 & .559 & .548 & .560 & .556 & .566 & .554 & .557 & .554 & .550 & .549 & .555 & 59.0\% \\
	CNN3, mine-4/4 & .628 & .619 & .635 & .632 & .639 & .625 & .636 & .631 & .624 & .626 & .630 & 80.5\% \\
	CNN3, mine 8/8 & .667 & .658 & .669 & .667 & .678 & .659 & .672 & .667 & .662 & .666 & {\bf .667} & {\bf 91.1\%} \\
	\bottomrule
\end{tabular}
\caption{Generalized results in Precision-Recall. Models trained over LY+YOS and tested on ND.}
\label{tbl:appendix:5-pr1}
\end{table}

\begin{table}[!h]
\centering
\def\arraystretch{1.1}
\setlength{\tabcolsep}{4pt}
\begin{tabular}{ccccccccccccc}
	\toprule
	& \multicolumn{10}{c}{{\bf Precision-Recall AUC, Train: LY+ND, Test: YOS (10 folds)}} & \\
	\hline
	Model & F1 & F2 & F3 & F4 & F5 & F6 & F7 & F8 & F9 & F10 & Avg. & Increase \\
	\hline \hline
	SIFT           & .428 & .419 & .413 & .416 & .414 & .427 & .429 & .442 & .432 & .430 & .425 & --- \\
	CNN3, mine-1/2 & .381 & .385 & .367 & .386 & .366 & .390 & .393 & .401 & .383 & .376 & .383 & -9.9\% \\
	CNN3, mine-2/2 & .388 & .395 & .377 & .393 & .376 & .397 & .401 & .405 & .388 & .381 & .390 & -8.2\% \\
	CNN3, mine-4/4 & .502 & .504 & .483 & .509 & .485 & .515 & .513 & .516 & .499 & .489 & .502 & 18.1\% \\
	CNN3, mine-8/8 & .547 & .547 & .528 & .551 & .528 & .559 & .556 & .561 & .546 & .530 & {\bf .545} & {\bf 28.2}\% \\
	\bottomrule
\end{tabular}
\caption{Generalized results in Precision-Recall. Models trained over LY+ND and tested on YOS.}
\label{tbl:appendix:5-pr2}
\end{table}

\begin{table}[!h]
\centering
\def\arraystretch{1.1}
\setlength{\tabcolsep}{4pt}
\begin{tabular}{ccccccccccccc}
	\toprule
	& \multicolumn{10}{c}{{\bf Precision-Recall AUC, Train: YOS+ND, Test: LY (10 folds)}} & \\
	\hline
	Model & F1 & F2 & F3 & F4 & F5 & F6 & F7 & F8 & F9 & F10 & Avg. & Increase \\
		\hline \hline
	SIFT           & .223 & .226 & .229 & .228 & .226 & .222 & .233 & .235 & .219 & .223 & .226 & --- \\
	CNN3, mine-1/2 & .460 & .464 & .464 & .460 & .454 & .452 & .462 & .463 & .462 & .456 & .460 & 103.5\% \\
	CNN3, mine-2/2 & .482 & .487 & .490 & .485 & .478 & .472 & .484 & .488 & .486 & .478 & .483 & 113.7\% \\
	CNN3, mine-4/4 & .564 & .566 & .569 & .562 & .560 & .557 & .564 & .567 & .570 & .562 & .564 & 149.6\% \\
	CNN3, mine-8/8 & .607 & .611 & .610 & .604 & .603 & .604 & .606 & .615 & .612 & .608 & {\bf .608} & {\bf 169.0}\% \\
	\bottomrule
\end{tabular}
\caption{Generalized results in Precision-Recall. Models trained over LY+ND and tested on YOS.}
\label{tbl:appendix:5-pr3}
\end{table}

\begin{table}[!h]
\centering
\def\arraystretch{1.1}
\setlength{\tabcolsep}{4pt}
\begin{tabular}{ccccccccccccc}
	\toprule
	& \multicolumn{10}{c}{{\bf ROC AUC, Train: LY+YOS, Test: ND (10 folds)}} & \\
	\hline
	Model & F1 & F2 & F3 & F4 & F5 & F6 & F7 & F8 & F9 & F10 & Avg. \\
	\hline \hline
	SIFT           & .956 & .954 & .955 & .958 & .957 & .955 & .955 & .955 & .956 & .955 & .956 \\
	CNN3, mine-1/2 & .979 & .978 & .979 & .982 & .980 & .978 & .981 & .981 & .979 & .979 & .980 \\
	CNN3, mine-2/2 & .981 & .980 & .981 & .983 & .982 & .980 & .983 & .982 & .981 & .981 & .981 \\
	CNN3, mine-4/4 & .985 & .984 & .985 & .987 & .986 & .985 & .988 & .986 & .985 & .985 & .986 \\
	CNN3, mine-8/8 & .986 & .985 & .986 & .988 & .987 & .986 & .989 & .986 & .986 & .986 & {\bf .987} \\
	\bottomrule
\end{tabular}
\caption{Generalized results in ROC. Models trained over LY+YOS and tested on ND.}
\label{tbl:appendix:5-roc1}
\end{table}

\begin{table}[!h]
\centering
\def\arraystretch{1.1}
\setlength{\tabcolsep}{4pt}
\begin{tabular}{ccccccccccccc}
	\toprule
	& \multicolumn{10}{c}{{\bf ROC AUC, Train: LY+ND, Test: YOS (10 folds)}} & \\
	\hline
	Model & F1 & F2 & F3 & F4 & F5 & F6 & F7 & F8 & F9 & F10 & Avg. \\
	\hline \hline
SIFT           & .949 & .947 & .948 & .949 & .949 & .950 & .949 & .950 & .950 & .950 & .949 \\
CNN3, mine-1/2 & .956 & .953 & .955 & .956 & .957 & .958 & .957 & .957 & .958 & .957 & .956 \\
CNN3, mine-2/2 & .958 & .955 & .957 & .958 & .959 & .959 & .958 & .959 & .960 & .958 & .958 \\
CNN3, mine-4/4 & .971 & .969 & .971 & .971 & .973 & .973 & .972 & .972 & .973 & .971 & .972 \\
CNN3, mine-8/8 & .974 & .972 & .975 & .974 & .976 & .975 & .975 & .975 & .976 & .974 & {\bf .975} \\
	\bottomrule
\end{tabular}
\caption{Generalized results in ROC. Models trained over LY+ND and tested on YOS.}
\label{tbl:appendix:5-roc2}
\end{table}

\begin{table}[!h]
\centering
\def\arraystretch{1.1}
\setlength{\tabcolsep}{4pt}
\begin{tabular}{ccccccccccccc}
	\toprule
	& \multicolumn{10}{c}{{\bf ROC AUC, Train: YOS+ND, Test: LY (10 folds)}} & \\
	\hline
	Model & F1 & F2 & F3 & F4 & F5 & F6 & F7 & F8 & F9 & F10 & Avg. \\
		\hline \hline
	SIFT           & .938 & .939 & .936 & .938 & .933 & .935 & .936 & .938 & .937 & .936 & .937 \\
	CNN3, mine-1/2 & .973 & .973 & .973 & .971 & .972 & .972 & .972 & .973 & .973 & .972 & .972 \\
	CNN3, mine-2/2 & .975 & .976 & .975 & .974 & .976 & .975 & .974 & .976 & .976 & .974 & .975 \\
	CNN3, mine-4/4 & .980 & .980 & .980 & .979 & .980 & .980 & .979 & .982 & .981 & .979 & .980 \\
	CNN3, mine-8/8 & .983 & .983 & .983 & .981 & .983 & .982 & .982 & .984 & .983 & .982 & {\bf .982} \\
	\bottomrule
\end{tabular}
\caption{Generalized results in ROC. Models trained over LY+ND and tested on YOS.}
\label{tbl:appendix:5-roc3}
\end{table}

\begin{table}[!h]
\centering
\def\arraystretch{1.1}
\setlength{\tabcolsep}{4pt}
\begin{tabular}{ccccccccccccc}
	\toprule
	& \multicolumn{10}{c}{{\bf CMC AUC, Train: LY+YOS, Test: ND (10 folds)}} & \\
	\hline
	Model & F1 & F2 & F3 & F4 & F5 & F6 & F7 & F8 & F9 & F10 & Avg. \\
	\hline \hline
	SIFT           & .964 & .962 & .963 & .966 & .965 & .963 & .964 & .963 & .964 & .962 & .963 \\
	CNN3, mine-1/2 & .982 & .981 & .981 & .984 & .982 & .981 & .983 & .983 & .982 & .981 & .982 \\
	CNN3, mine-2/2 & .983 & .982 & .983 & .985 & .984 & .982 & .985 & .984 & .984 & .983 & .984 \\
	CNN3, mine-4/4 & .987 & .987 & .987 & .989 & .988 & .987 & .989 & .988 & .987 & .987 & .988 \\
	CNN3, mine-8/8 & .988 & .988 & .988 & .990 & .989 & .988 & .990 & .989 & .989 & .989 & {\bf .989} \\
	\bottomrule
\end{tabular}
\caption{Generalized results in CMC. Models trained over LY+YOS and tested on ND.}
\label{tbl:appendix:5-cmc1}
\end{table}

\begin{table}[!h]
\centering
\def\arraystretch{1.1}
\setlength{\tabcolsep}{4pt}
\begin{tabular}{ccccccccccccc}
	\toprule
	& \multicolumn{10}{c}{{\bf CMC AUC, Train: LY+ND, Test: YOS (10 folds)}} & \\
	\hline
	Model & F1 & F2 & F3 & F4 & F5 & F6 & F7 & F8 & F9 & F10 & Avg. \\
	\hline \hline
	SIFT           & .956 & .955 & .956 & .956 & .956 & .958 & .956 & .957 & .956 & .958 & .956 \\
	CNN3, mine-1/2 & .964 & .962 & .963 & .964 & .965 & .966 & .964 & .965 & .966 & .966 & .965 \\
	CNN3, mine-2/2 & .966 & .965 & .966 & .967 & .968 & .968 & .967 & .968 & .969 & .968 & .967 \\
	CNN3, mine-4/4 & .977 & .976 & .978 & .978 & .980 & .980 & .977 & .979 & .980 & .980 & .978 \\
	CNN3, mine-8/8 & .980 & .979 & .981 & .981 & .982 & .982 & .980 & .982 & .982 & .982 & {\bf .981} \\
	\bottomrule
\end{tabular}
\caption{Generalized results in CMC. Models trained over LY+ND and tested on YOS.}
\label{tbl:appendix:5-cmc2}
\end{table}

\begin{table}[!h]
\centering
\def\arraystretch{1.1}
\setlength{\tabcolsep}{4pt}
\begin{tabular}{ccccccccccccc}
	\toprule
	& \multicolumn{10}{c}{{\bf CMC AUC, Train: YOS+ND, Test: LY (10 folds)}} & \\
	\hline
	Model & F1 & F2 & F3 & F4 & F5 & F6 & F7 & F8 & F9 & F10 & Avg. \\
		\hline \hline
	SIFT           & .948 & .949 & .947 & .948 & .945 & .945 & .948 & .949 & .948 & .947 & .948 \\
	CNN3, mine-1/2 & .976 & .975 & .976 & .974 & .975 & .975 & .975 & .977 & .976 & .975 & .975 \\
	CNN3, mine-2/2 & .978 & .979 & .978 & .977 & .978 & .978 & .978 & .979 & .979 & .977 & .978 \\
	CNN3, mine-4/4 & .983 & .983 & .983 & .982 & .983 & .982 & .982 & .984 & .984 & .982 & .983 \\
	CNN3, mine-8/8 & .985 & .985 & .985 & .984 & .985 & .984 & .985 & .986 & .986 & .985 & {\bf .985} \\
	\bottomrule
\end{tabular}
\caption{Generalized results in CMC. Models trained over LY+ND and tested on YOS.}
\label{tbl:appendix:5-cmc3}
\end{table}

\clearpage

\end{document}